\documentclass[10pt,twocolumn,letterpaper]{article}

\usepackage[nocompress]{cite}
\usepackage{iccv}
\usepackage{times}
\usepackage{epsfig}
\usepackage{graphicx}
\usepackage{amsmath}
\usepackage{amssymb}
\usepackage[noblocks]{authblk}

\makeatletter
\renewcommand\AB@affilsepx{ --- \protect\Affilfont}
\makeatother

\usepackage[breaklinks=true,bookmarks=false]{hyperref}

\iccvfinalcopy 

\def\iccvPaperID{16} 

\ificcvfinal\fi
\begin{document}

\title{Integral Curvature Representation and Matching Algorithms\\ for
Identification of Dolphins and Whales}

\renewcommand\Authands{, }
\author[1]{Hendrik J. Weideman}
\author[1]{Zachary M. Jablons}
\author[2]{Jason Holmberg}
\author[3]{Kiirsten Flynn}
\author[3]{John Calambokidis}
\author[4]{Reny B. Tyson}
\author[4]{Jason B. Allen}
\author[4]{Randall S. Wells}
\author[5]{Krista Hupman}
\author[6]{Kim Urian}
\author[1]{Charles V. Stewart}
\affil[1]{Rensselaer Polytechnic Institute}
\affil[2]{WildMe}
\affil[3]{Cascadia Research Collective}
\affil[4]{Chicago Zoological Society's Sarasota Dolphin Research Program, c/o
Mote Marine Laboratory}
\affil[5]{{Massey University}}
\makeatletter
\renewcommand\AB@affilsepx{\\\protect\Affilfont}
\makeatother
\affil[6]{Duke University Marine Laboratory}
\affil[1]{\tt\small {\{weideh,jabloz2,stewart\}@rpi.edu}}
\makeatletter
\renewcommand\AB@affilsepx{ \protect\Affilfont}
\makeatother
\affil[2]{\tt\small {holmbergius@gmail.com}}
\affil[3]{\tt\small {\{kflynn,calambokidis\}@cascadiaresearch.org}}
\affil[4]{\tt\small {\{rtyson,allenjb,rwells\}@mote.org}}
\affil[5]{\tt\small {k.rankmore@massey.ac.nz}}
\affil[6]{\tt\small {kim.urian@gmail.com}}

\maketitle
\thispagestyle{empty}

\begin{abstract}
  We address the problem of identifying individual cetaceans from images
  showing the trailing edge of their fins.  Given the trailing edge from an
  unknown individual, we produce a ranking of known individuals from a
  database.  The nicks and notches along the trailing edge define an
  individual's unique signature.  We define a representation based on integral
  curvature that is robust to changes in viewpoint and pose, and captures the
  pattern of nicks and notches in a local neighborhood at multiple scales.  We
  explore two ranking methods that use this representation.  The first uses a
  dynamic programming time-warping algorithm to align two representations, and
  interprets the alignment cost as a measure of similarity.  This algorithm
  also exploits learned spatial weights to downweight matches from regions of
  unstable curvature.  The second interprets the representation as a feature
  descriptor.  Feature keypoints are defined at the local extrema of the
  representation.  Descriptors for the set of known individuals are stored in a
  tree structure, which allows us to perform queries given the descriptors from
  an unknown trailing edge.  We evaluate the top-$k$ accuracy on two real-world
  datasets to demonstrate the effectiveness of the curvature representation,
  achieving top-$1$ accuracy scores of approximately $95\%$ and $80\%$ for
  bottlenose dolphins and humpback whales, respectively.
\end{abstract}

\section{Introduction}
\begin{figure}[t]
\begin{center}
  \includegraphics[width=0.45\linewidth]{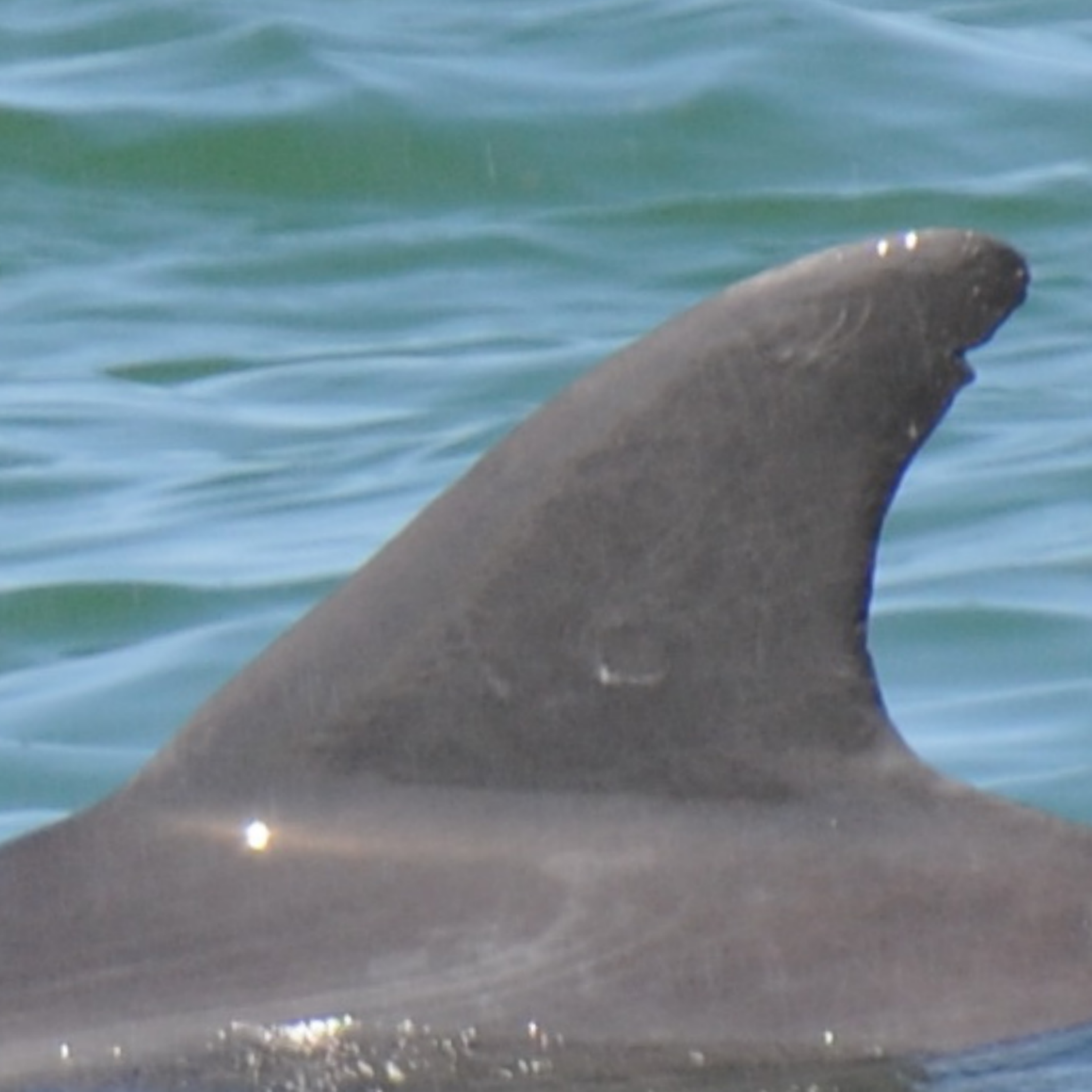}
  \includegraphics[width=0.45\linewidth]{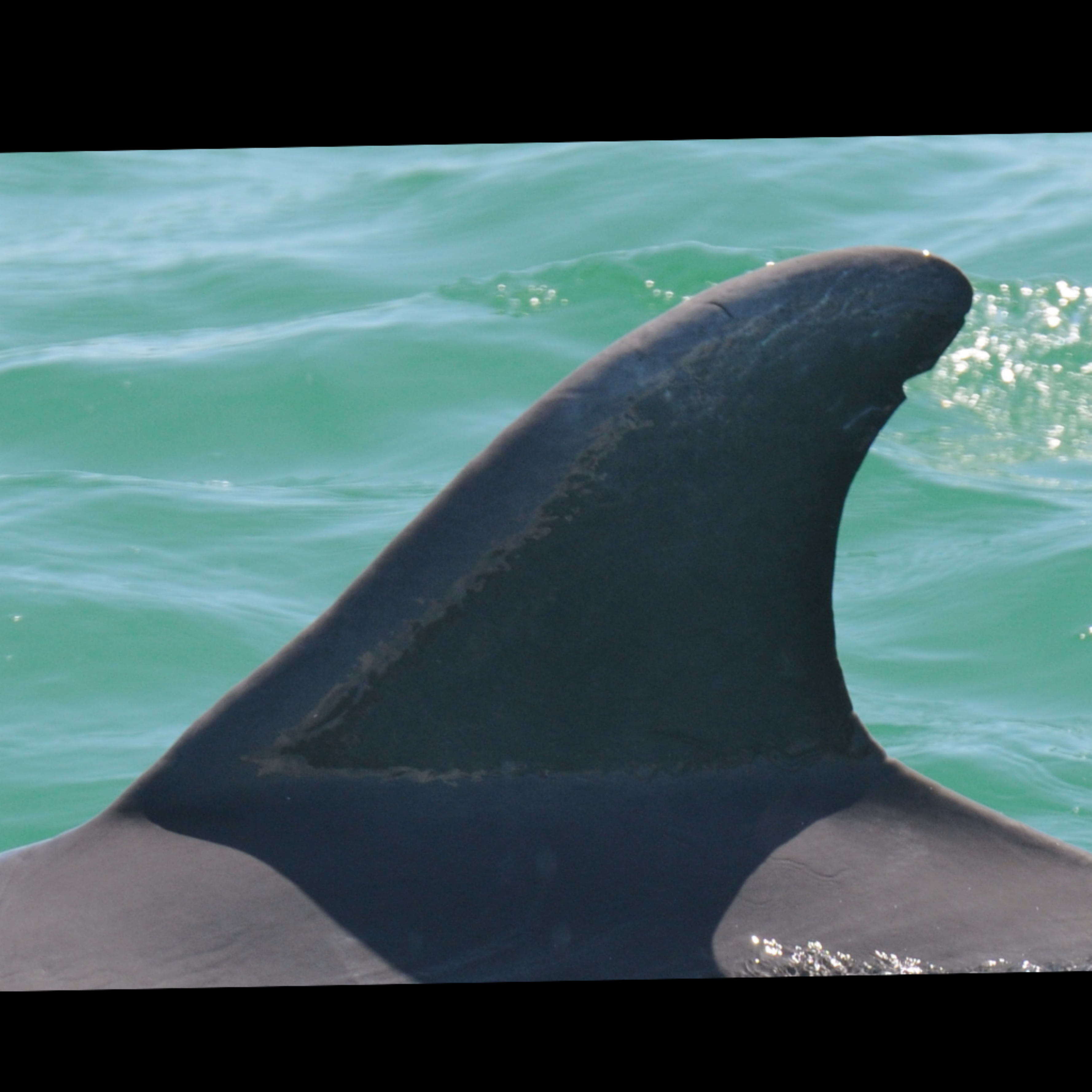}\\
  \includegraphics[width=0.45\linewidth]{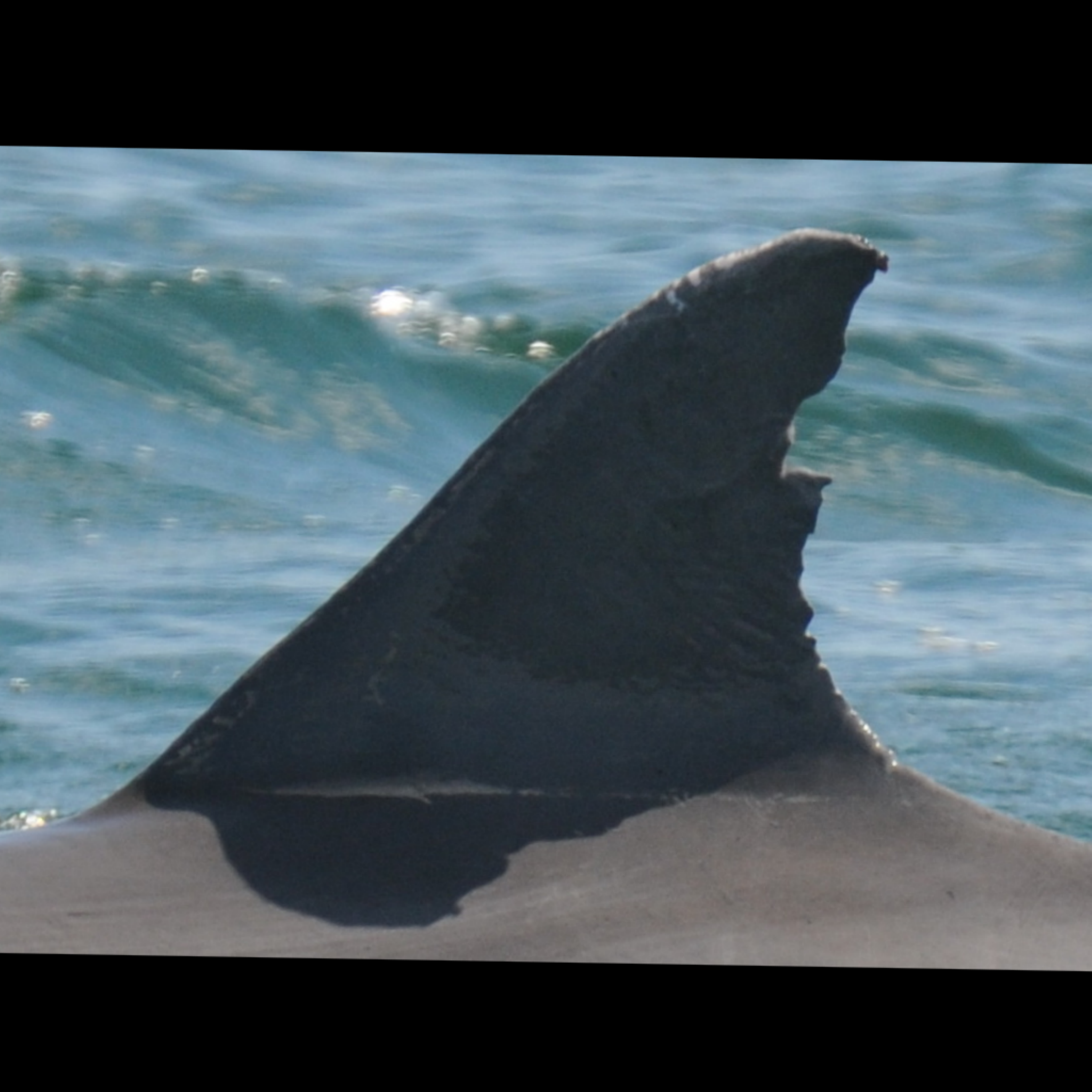}
  \includegraphics[width=0.45\linewidth]{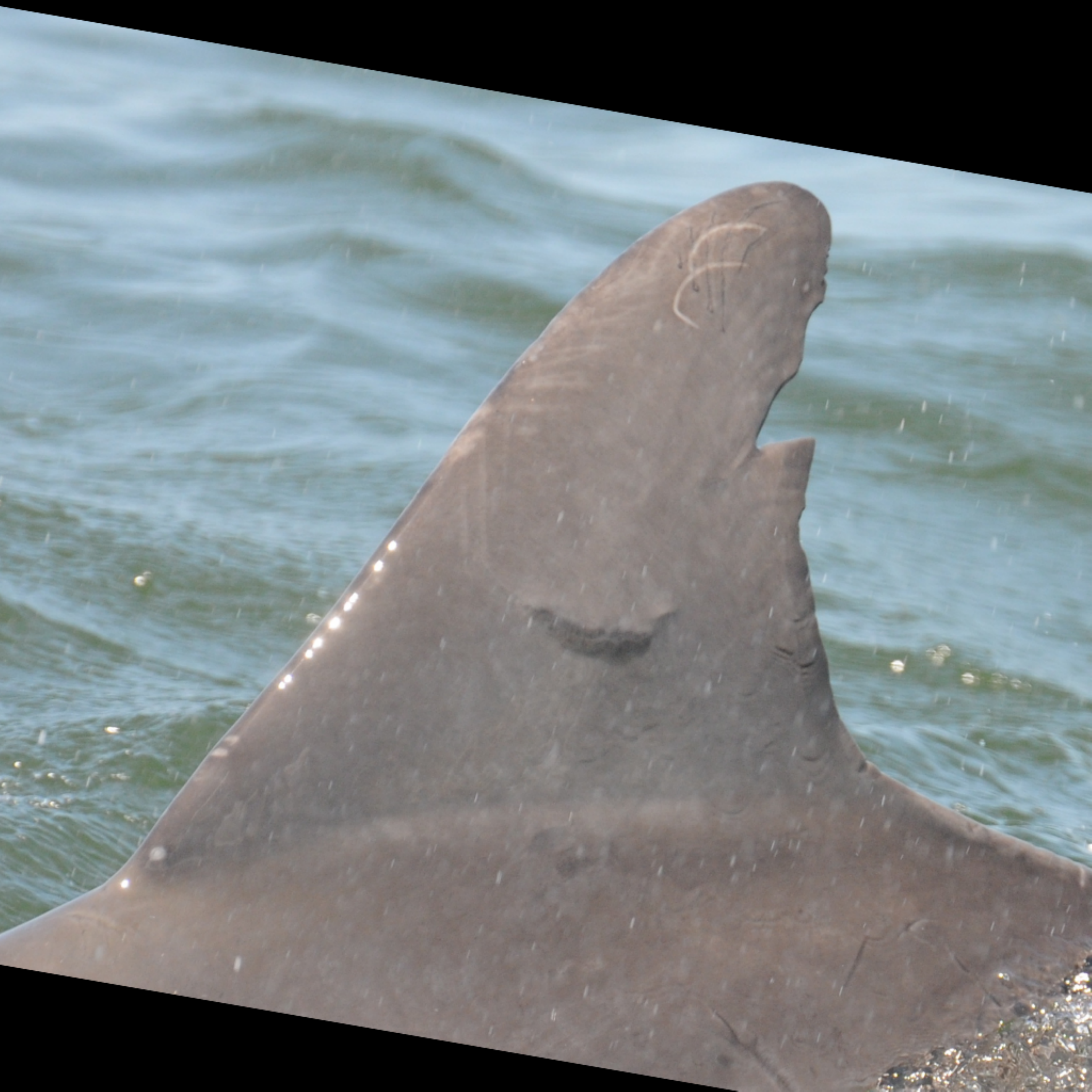}
\end{center}
\vspace{-0.1in}
\caption{Example images of dorsal fins from the \emph{Bottlenose} dolphin
  dataset.  Although the fins in each row may appear similar, they are from
  distinct individuals.  Note that the identifying information in each fin
  comes from one or two large markings. Compare this to the case
  for the \emph{Humpback} dataset, where the identifying information is spread along the entire contour.
}
\vspace{-0.1in}
\label{fig:bottlenose}
\end{figure}

\begin{figure}[t]
\begin{center}
  \includegraphics[width=0.9\linewidth]{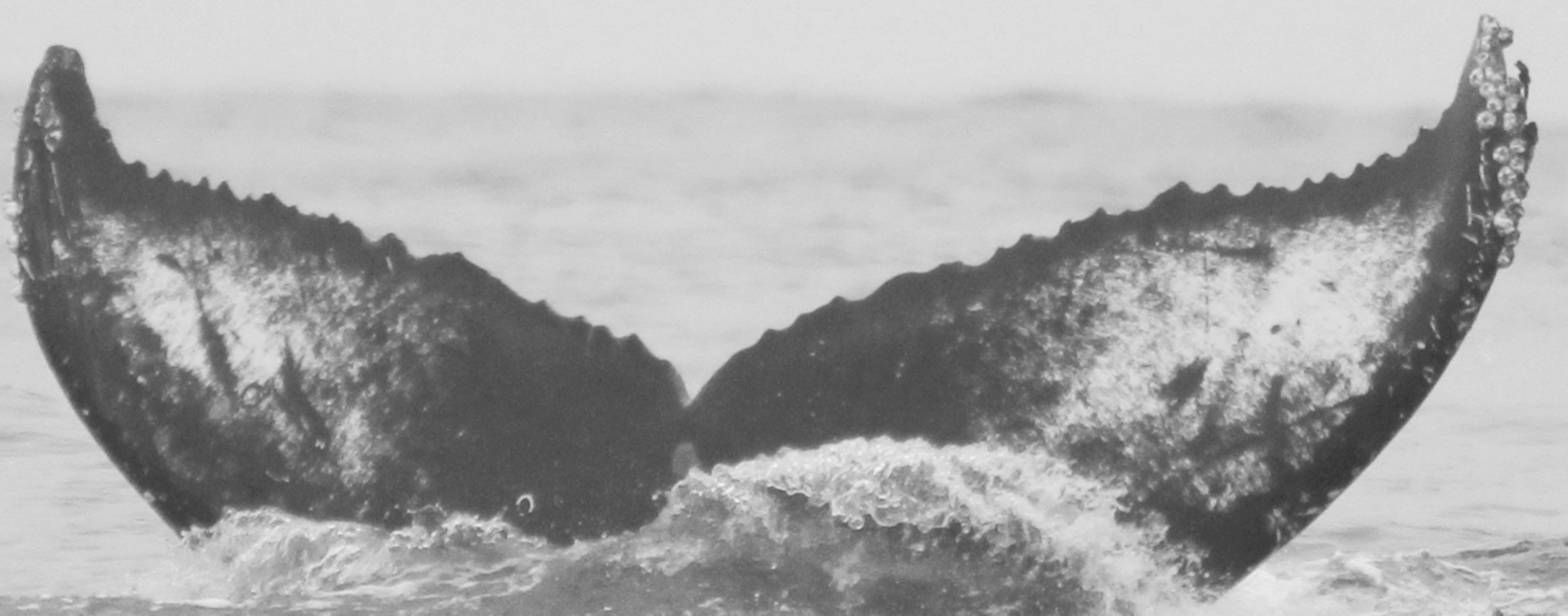}\\
  \includegraphics[width=0.9\linewidth]{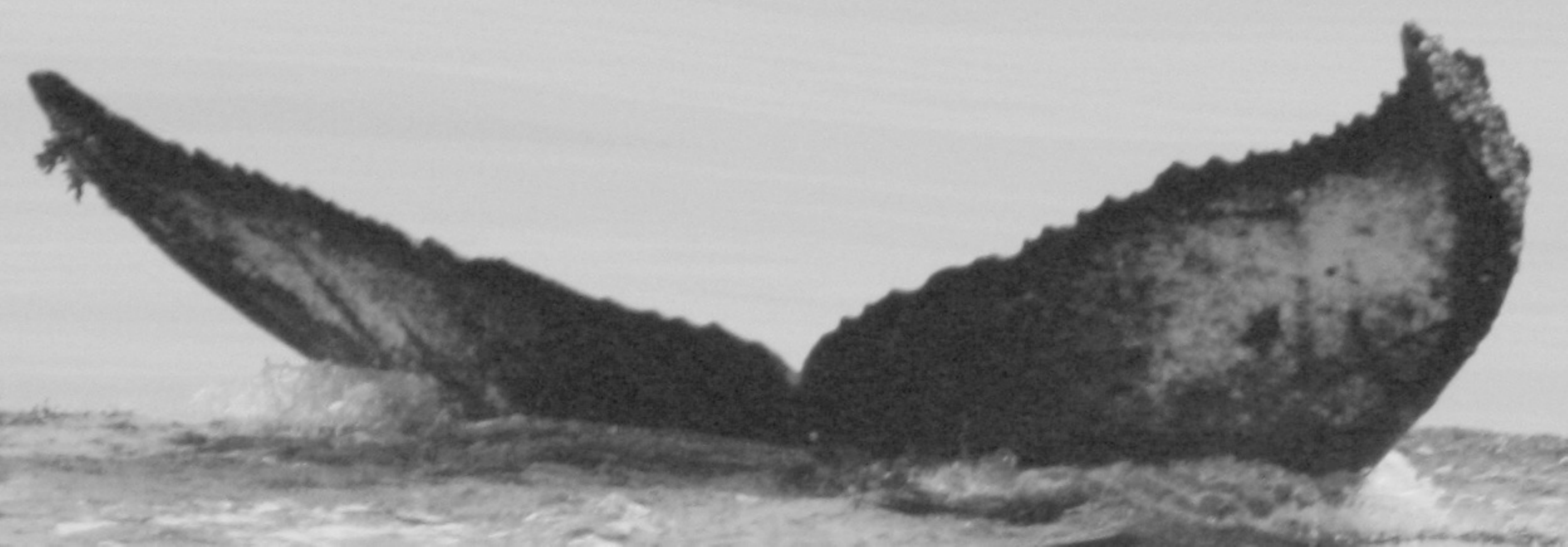}\\
  \includegraphics[width=0.9\linewidth]{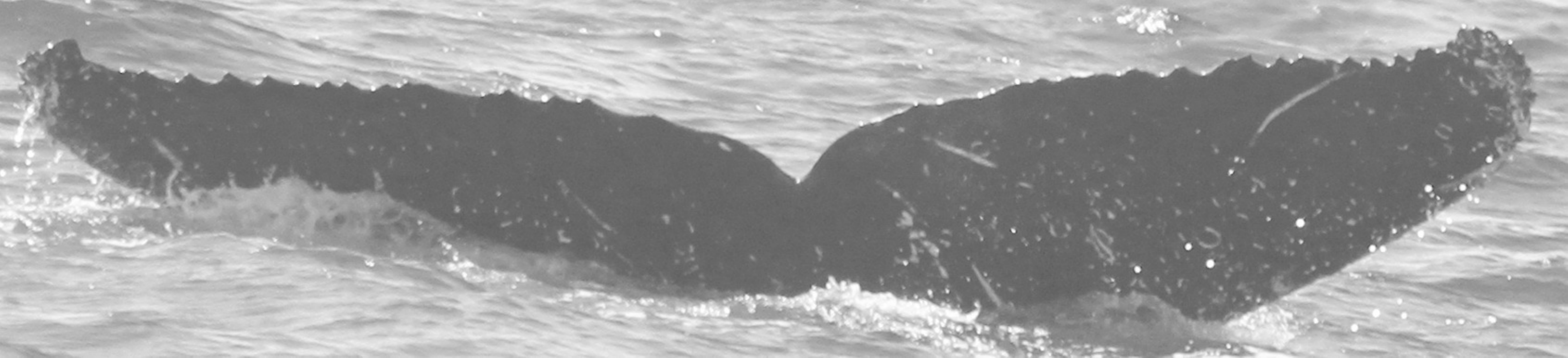}\\
  \includegraphics[width=0.9\linewidth]{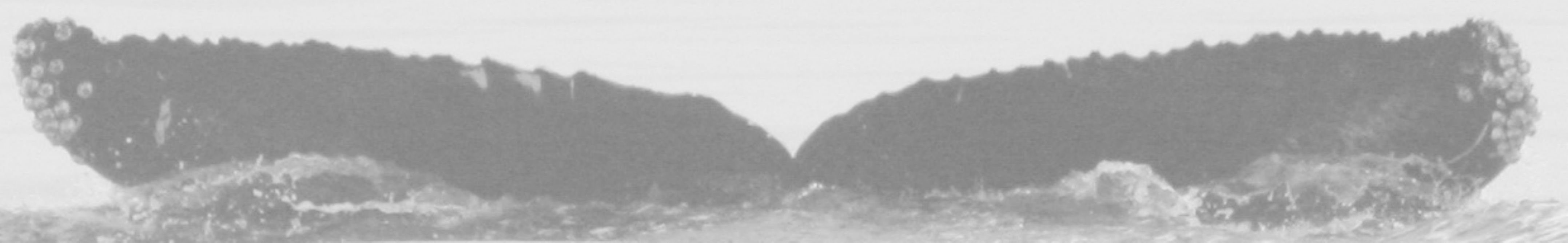}\\
\end{center}
\vspace{-0.1in}
\caption{Example images of flukes from the \emph{Humpback} whale dataset.  The
  two upper images are from the same individual, while the two lower images are
  from distinct individuals.  Note that, unlike for the \emph{Bottlenose}
  dataset, the information necessary for identification is spread along the
  entire length of the fluke trailing edge.
}
\vspace{-0.1in}
\label{fig:humpback}
\end{figure}

We address the problem of identifying individual cetaceans from images of their
fins --- dorsal fins for bottlenose dolphins, and flukes for humpback whales.
Fitting into the broad domain of contour-based recognition~\cite{Leibe03}, the fin instance recognition
problem is particularly challenging, as illustrated in
Figures~\ref{fig:bottlenose}~and~\ref{fig:humpback}.

Our hypothesis is that the information necessary to distinguish between the
outlines of fins from distinct individuals is encoded in local measures of
integral curvature~\cite{Jablons16}.  Differential curvature measures have been applied to a
variety of recognition problems, but these approaches are sensitive to
noise~\cite{Pottmann09}.  Integral curvature, which produces more stable
measurements, has been applied to category recognition problems like
identifying leaf species~\cite{Kumar12}.

In this paper, we propose novel combinations of integral curvature
representation for extracted outline contours and two matching algorithms for
identifying individuals.  The first interprets the curvature representation as
a sequence, and defines the similarity between two representations as the cost
of warping one sequence onto the other.  The second treats subsections of the
representation as feature descriptors, and matches them using approximate
nearest neighbors.  Each takes a query image and produces a ranking of known
individuals from a database.  We also introduce a method for learning spatial
weights that describe the relative importance of points along the trailing
edge, which enables the matching algorithm to assign higher value to the most
distinguishing areas of the fin contours.

In cetacean research, identifying individuals observed during a survey is a
fundamental part of studying populations.  Traditionally, techniques such as
tagging or branding are used to make identification easier.  Not only do these
require specialized equipment and training, but they also tend to be invasive,
requiring catch-and-release of wild animals.  As a less invasive alternative,
photo identification requires no direct contact with the animals. Instead,
researchers use images of long-lasting markings, such as nicks, notches, or
scars, to track individuals over time~\cite{Wells90, Hammond90}.

Photo identification presents its own challenges. The images showing distinct
individuals observed during a study need to be matched against a database
containing images of known individuals.  For large cetacean populations
covering a large geographical area and monitored over a long time period, this
can be very time-consuming when done using manual methods.  Additionally, parts
of the trailing edge required for identification are often occluded by water or
viewpoint, requiring the use of multiple images per individual.  Even when the
trailing edge contours are consistently visible, direct matching between
trailing edges from different images is problematic.  Because the images are
captured ``in the wild'', animals occur in a wide variety of poses and are
photographed from varying viewpoints.  Even for a small change in viewpoint,
out-of-plane rotations can lead to difficulty in matching nicks and notches.

By presenting likely matches for a query individual to the user in order of
similarity, manual identification may be substantially accelerated by reducing
(on average) the number of individuals to compare per query.

\subsection{Time-Warping Sequence Alignment} 
It is possible to treat the curvature representations of the trailing edges as
vectors and compute their similarity using any vector norm.  The start and
endpoints of fins are ambiguous, and pose and viewpoint stretch some sections
and foreshorten others, so we need to account for nicks and notches occurring
at differing locations along the trailing edge.  We use a dynamic programming
time-warping algorithm~\cite{Sakoe78} that computes the alignment cost of two
representations.  Rather than a one-to-one matching, as with a vector norm,
this allows many points from one representation to match to one point in the
other, and vice versa.  This warps one representation onto the other, where the
alignment cost is the sum of errors for local correspondences~\cite{Jablons16}.

\subsection{Descriptor Indexing} 
Considering that the nicks and notches used for identification are often
sparsely distributed along the trailing edge, and that points in between offer
little value for identification, it seems desirable to use only the former when
performing an identification.  To do this, we compute local extrema in the
trailing edge representation, i.e., points corresponding to regions of high
curvature in the original trailing edge, and use these as feature
keypoints~\cite{Hughes16}.  Between these keypoints, we extract feature
descriptors, resample them to a fixed length, and normalize using the Euclidean
norm.  All individuals from the database are stored in a tree-like structure,
after which the most likely candidate matches for a given trailing edge are
computed using the local naive Bayes nearest neighbor classifier~\cite{McCann12}.

\subsection{Contributions}
The primary contributions of this work are that we (a) develop an integral
curvature measure to represent trailing edge contours in such a way that the
representation is robust to changes in viewpoint and pose that make direct
comparison difficult, (b) propose integration of these measures with
time-warping alignment and descriptor indexing algorithms as two approaches for
ranking potential matches, (c) develop a learning algorithm to weight sections
of the trailing edge contour, and (d) produce results using this representation
and these algorithms on real-world datasets from active cetacea research groups
to confirm their efficacy.

\section{Related Work}
The curvature computed at points along a contour is often used to represent
shape information~\cite{Monroy11, Kumar12, Fischer14}.  For differential
curvature, this is defined as the change of the angle of the normal vector
along the length of the contour~\cite{Fischer14}.  This representation is
sensitive to noise, however, and instead we use integral curvature.  A fixed
shape is placed at points along the contour, while measuring the area of the
intersection of the shape with the contour~\cite{Pottmann09}.  Using integral
curvature to capture shape information is a key part of the Leafsnap
system~\cite{Kumar12}, which classifies leaves by using curvature histograms at
multiple scales as shape features.  We briefly compare to Leafsnap in
Section~\ref{sec:experiments}.

In terms of identifying dolphins from their dorsal fins, most notable is
DARWIN~\cite{Stanley95}.  The key idea behind DARWIN is to account for changes
in viewpoint by computing a transformation between two fins to align them, and
using the resulting sum of squared distances to define a similarity
score~\cite{Stewman06}.  As pointed out in~\cite{Gilman13}, a fundamental
problem with this approach is that fins from distinct individuals are unlikely
to align correctly, with the result that similarity scores cannot be compared
reliably to produce a ranking.

Another system, Finscan~\cite{Hillman02}, frames the problem of comparing two
dorsal fins as a string matching problem~\cite{Araabi00}.  In this setting,
they compute a low-level string representation of the trailing edge curvature.
Because the curvature function defined in terms of derivatives is typically
noisy, they refine this representation to a high-level string
representation.  To compute the similarity between two string representations,
they use a linear time-warping algorithm.

In both DARWIN and Finscan, the extraction of the trailing edge is a
semi-automatic, interactive process where the user manually corrects an initial
estimate of the trailing edge.  The size of our datasets, as well as our goal
of scaling to real-world scenarios, makes comparison with these
systems impractical.  Instead, we design our representation and matching
algorithms to be robust against the occasional inaccurate trailing edge
extraction.

In~\cite{Hughes16}, the authors propose a trailing edge indexing algorithm and
apply it to great white sharks.  After defining keypoints for feature
extraction by convolving the contour with a Difference-of-Gaussian kernel, they
explore the use of both the Difference-of-Gaussian norm and the descriptor
from~\cite{Arandjelovic12}.  The descriptor indexing algorithm in our work is
similar, except that we instead use the curvature representation as a feature
descriptor.

A notable approach to the more general problem of ranking is the \emph{triplet
network}~\cite{Hoffer15, Wang14}, a natural extension of the \emph{Siamese
network}~\cite{Taigman14, Chopra05}.  A triplet network is a neural network that learns
a useful representation by minimizing the distance between the representations
of instances of the same class, while maximizing the distance between
representations from different classes.  One key advantage of this approach is
that the representation does not need to be designed by hand, rather, it is
learned from a large set of labeled training data.  The resulting
representation is then used to embed query instances into the same space, where
a ranking may be computed.  We briefly explored the use of triplet networks
using the original trailing edge contours as well as the curvature
representation, but found that our small datasets
would lead to overfitting.  Additionally, the parametric nature of these models
makes it difficult to apply them to new datasets~\cite{Yosinski14}, whereas we confirm in
Section~\ref{subsec:ranking-performance} that our approach may be applied
unchanged to an unseen dataset of the same species.

\section{Datasets and Preprocessing}
\label{section:datasets}
We use two real-world datasets provided by active research groups to evaluate
our approach.  The first dataset is provided by the Sarasota Dolphin Research
Program and is illustrated in Figure~\ref{fig:bottlenose}.  This dataset
contains $10,\!713$ images representing $401$ distinct bottlenose dolphins
(\emph{Tursiops truncatus}).  Researchers take photos of the dolphins
encountered at a particular time and place, and the best images of each
individual are separated into \emph{encounters}.  These images are cropped to
the dorsal fin and added to the dataset.  The second dataset is provided by the
Cascadia Research Collective and shown in Figure~\ref{fig:humpback}.  This
dataset contains $7,\!173$ images representing $3,\!572$ humpback whales
(\emph{Megaptera novaeangliae}).  Unlike the first dataset where each
individual appears in multiple images per encounter, here each individual
typically appears in only a single image per encounter.  

Given an image, a fully-convolutional neural network (FCNN)~\cite{Long15}
outputs the probability that each pixel is part of the trailing edge.  Anchor
points are computed, and a shortest-path algorithm selects pixels based
on costs determined by a combination of the FCNN and image gradients.
For dorsal fins, this includes a spatial transformer network~\cite{Jaderberg15}
that transforms the image such that the fin is approximately perpendicular to
the image plane.

\section{Individual Identification}

\subsection{Curvature Representation}
\label{subsec:curvature-representation}
\begin{figure}[t]
\begin{center}
  \includegraphics[width=0.9\linewidth]{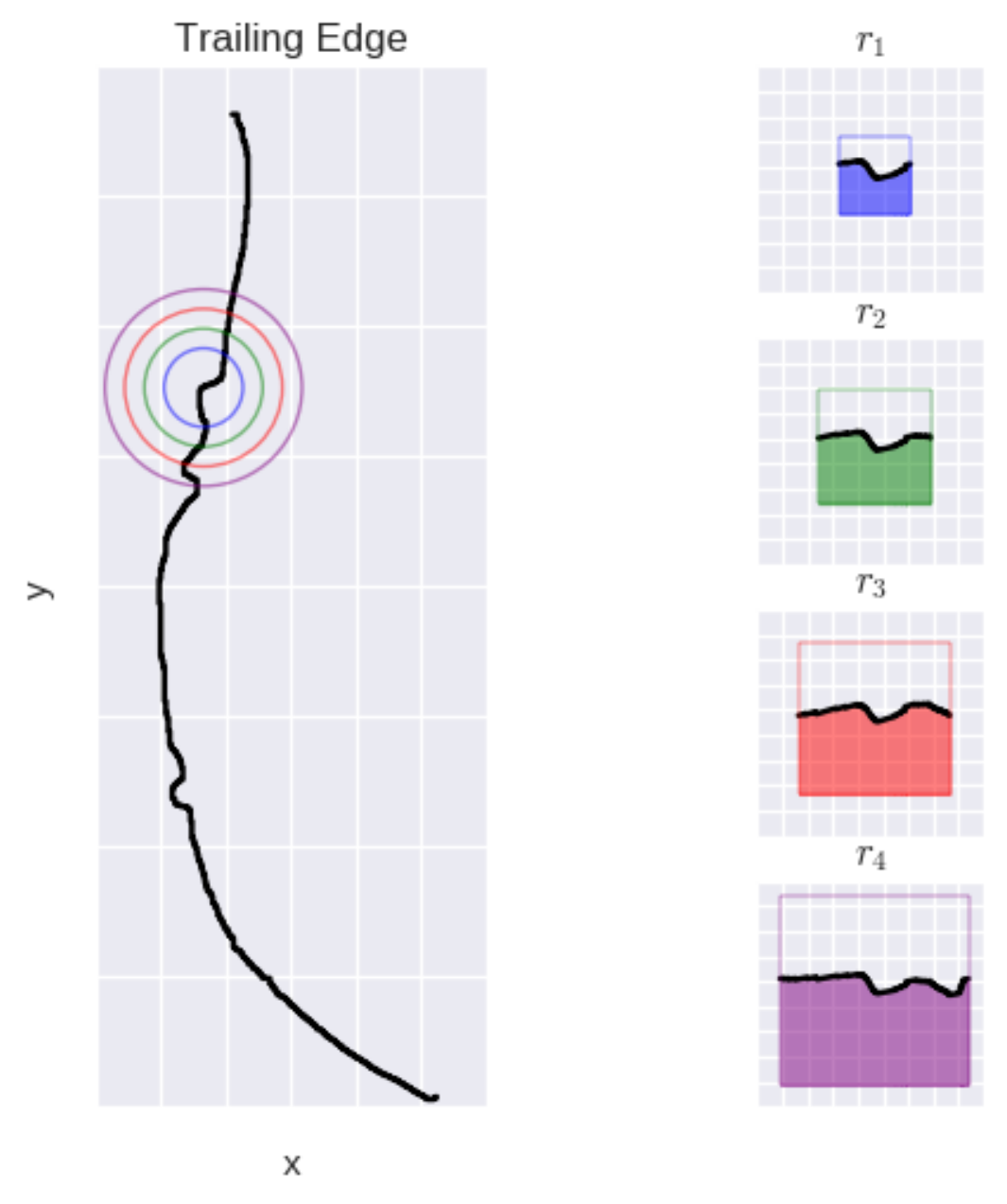}
\end{center}
\vspace{-0.1in}
\caption{For a given point $(x_j, y_j)$, the curve segments lying inside
  circles of radii $r_1, r_2, \ldots, r_m$ (left) are transformed to be
  horizontal (right).  The curvature at the point for a particular $r$ is then
  defined as the ratio of the area under the curve (shaded) to the area of a
  square of side length $2r$.
}
\vspace{-0.1in}
\label{fig:curv}
\end{figure}

Given a trailing edge contour represented as an ordered set of coordinates,
$\{(x_1, y_1), (x_2, y_2), \ldots, (x_n, y_n)\}$, we wish to represent this
contour such that it is robust to changes in viewpoint and pose.  For this we
use an integral curvature measure that captures local shape information at each
point along the trailing edge.  For a given point $(x_i, y_i)$ that lies on the
trailing edge, we place a circle of radius $r$ at the point and find all points
on the trailing edge that lie within this circle, i.e., $\mathbf{p}_i = \{(x_j,
y_j) \mid (x_i - x_j)^2 + (y_i - y_j)^2 \leq r^2\}$.  To describe a point
$(x_i, y_i)$  by its local curvature, we first orient the points $\mathbf{p}_i$
such that $\mathbf{p}_i(1)$ and $\mathbf{p}_i(n)$ lie on a horizontal line.
The coordinates of the points in $\mathbf{p}_i$ are then clipped to the
dimensions of an axis-aligned square with side length $2r$ centered at $(x_i,
y_i)$.  Using the bottom side of this square as the axis for the independent
variable, we use trapezoidal integration to approximate the area under the
curve defined by the discrete points $\mathbf{p}_i$.  We define the curvature
$c \in [0, 1]$ at this point as the ratio of the area under the curve to the
total area of the square, which implies that the curvature value for a straight
line is $c = 0.5$.  See Figure~\ref{fig:curv} for an illustration. By
computing this integral curvature measure at all points along the trailing
edge, we obtain the curvature representation of the trailing edge for a single
value of $r$.  To control the extent to which we capture local and global
information, we vary the radius of the circle placed at each point by choosing
multiple values of $r$ (typically four).  The result is a matrix $C$ of dimensions $m \times n$,
where $m$ is the number of values of $r$ that we choose and
$n$ is the number of points along the trailing edge.  The scalar $C_{ij} \in
[0, 1]$ is the curvature value for the $i$th value of $r$ at the $j$th point
along the trailing edge.

\begin{figure}[t]
\begin{center}
  \includegraphics[width=0.9\linewidth]{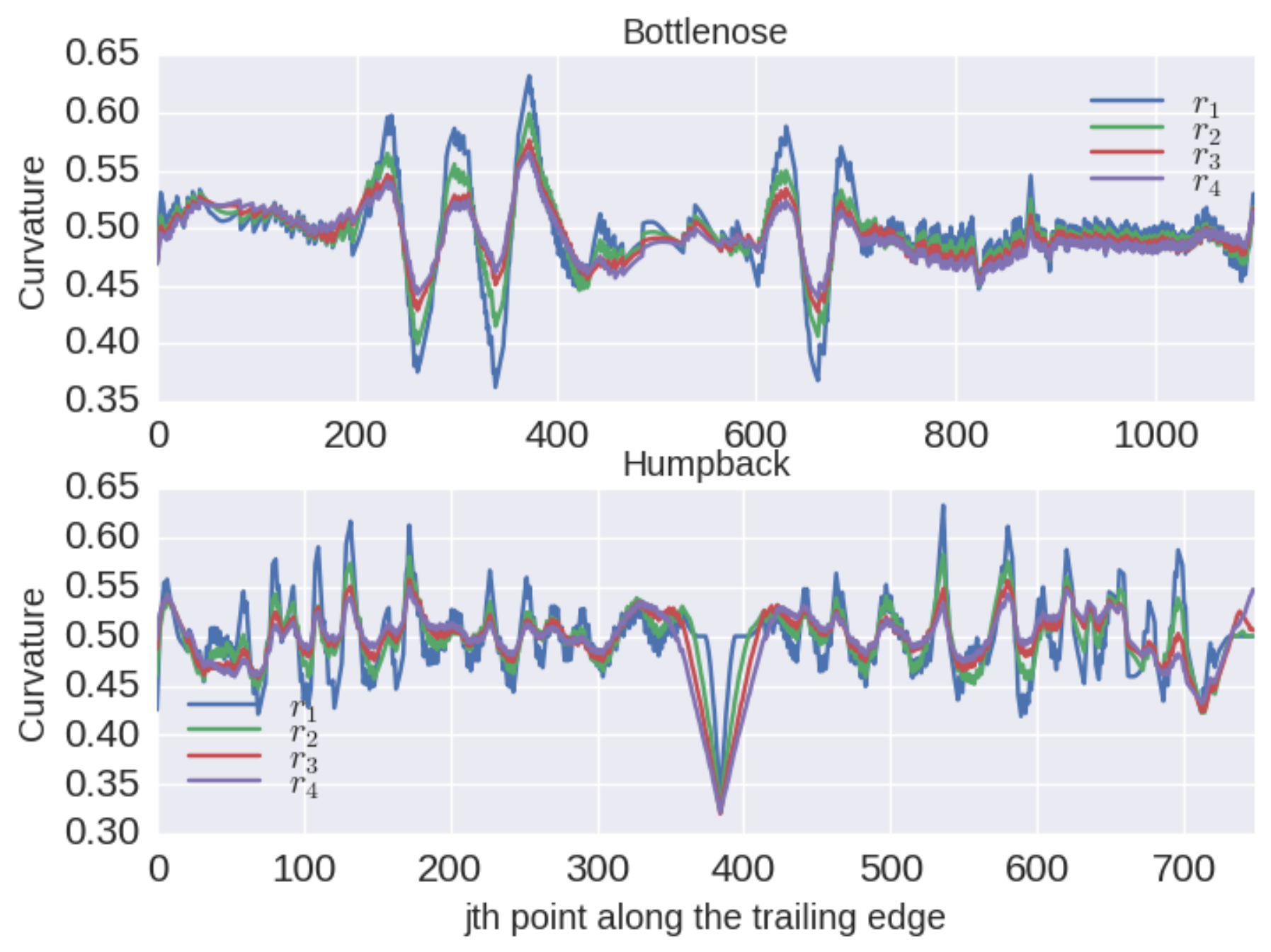}
\end{center}
\vspace{-0.1in}
\caption{Curvature representation of a dorsal fin from the \emph{Bottlenose}
  dataset (top) and fluke from the \emph{Humpback} dataset (bottom) computed for four different values of $r$.
}
\vspace{-0.1in}
\label{fig:curvature}
\end{figure}

\subsection{Ranking}
\label{subsec:ranking}
We explore two types of methods that use the curvature representation defined in
Section~\ref{subsec:curvature-representation} to
produce a ranking of known individuals given a query.
\subsubsection{Sequence Alignment}
\textbf{Dynamic Time-Warping.}
The first method for comparing representations that we explore interprets the
curvature representations as temporal sequences and computes an alignment cost
between them~\cite{Jablons16, Sakoe78}.  Given that the start and
endpoints of the trailing edges are not only ambiguous, but also often under
water, it is desirable to allow some degree of warping when computing
correspondences.  If we define $\mathbf{c}_i$ and $\mathbf{c}_j'$ as the $i$th
and $j$th columns (each a vector representing the curvature values at a point) of two
curvature representations $C$ and $C'$ of lengths $m$ and $m'$, respectively, then the total
alignment cost $c(m, m')$ is defined recursively as 
\begin{align}
  \mbox{\hspace{-2mm}}
  c(i, j) &= d(\mathbf{c}_i, \mathbf{c}_j') \nonumber \\
  &+ \min \{c(i-1, j), c(i, j-1),
  c(i-1, j-1)\},
\end{align}
where $d$ defines the distance between two points based on their curvature
values at multiple scales.  It is possible to use a simple vector norm, such as
$d(\mathbf{c}_i, \mathbf{c}_j') =
||\mathbf{c}_i-\mathbf{c}_j'||_2$.  

\textbf{Spatial Weights.}
The definition above, however, treats the contribution from correspondences
along the entire length of the trailing edge as equal.  We know that the end of
the trailing edge is often underwater for dorsal fins, and the tips often
cropped for flukes.  Ideally, we would thus like corresponding points from
these unstable regions to contribute less towards the total alignment cost.

To realize the above, we define a weight vector $\mathbf{w}$, where the
elements $w_1, w_2, \ldots, w_n$ describe the relative importance of each point
along the trailing edge.  In doing so, we define a more meaningful distance
function as
\vspace{-0.1in}
\begin{align}
  d(\mathbf{c}_i, \mathbf{c}_j' | \mathbf{w}) = w_iw_j||\mathbf{c}_i - \mathbf{c}_j'||_2,
\end{align}
where the product $w_iw_j$ scales the contribution of each correspondence to
the total alignment cost based on the relative importance of the points.

\textbf{Learning the Weights.}
To determine suitable values for the elements of $\mathbf{w}$, we frame the
problem as an unconstrained optimization problem where we maximize the top-$k$
score (the fraction of times the correct individual appears in the first $k$
entries of the ranking) over a training set.  For the training set we use the
images in the database, and take the images from a single encounter for each individual
to be used as queries.  The separation of images into database and queries is
described in Section~\ref{subsec:defining-queries}.  

To avoid overfitting, rather than learn all the
elements of $\mathbf{w}$, we reduce the number of parameters by expressing
$\mathbf{w}$ as a linear combination of the Bernstein polynomials~\cite{Lorentz12} of degree
$n$ evaluated at uniformly spaced points between $0$ and $1$.  The linear
combination of Bernstein polynomials determined by coefficients $\mathbf{c}$ is defined as
\begin{align}
  B_n(x) = \sum\limits_{i=0}^{n}c_ib_{i,n}(x),
\end{align}
where
\begin{align}
  b_{i,n}(x) = {n\choose i}x^i(1 - x)^{n - i},\quad i = 0, 1, \ldots, n.
\end{align}
These polynomials have two particular properties that are desirable for our
application, namely that~(a)~they are positive between $0$ and $1$, i.e.,
$b_{i,n}(x) \geq 0$ for $x \in [0, 1]$, and~(b)~they form a partition of unity, i.e., $\sum\limits_{i=0}^{n}b_{i,n}(x) = 1$.

\begin{figure}[t]
\begin{center}
  \includegraphics[width=0.9\linewidth]{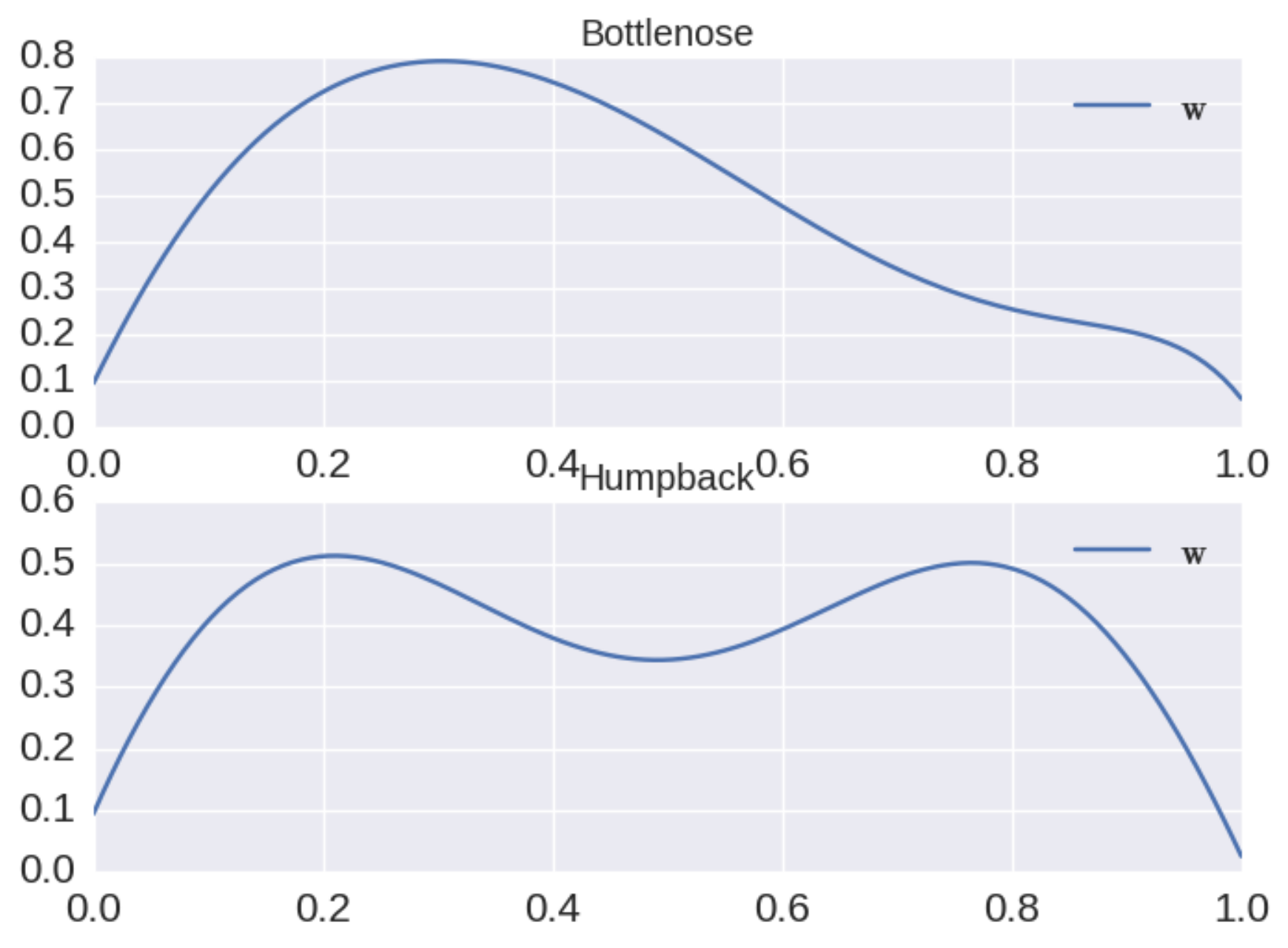}
\end{center}
\vspace{-0.1in}
\caption{The spatial weight vector $\mathbf{w}$ obtained by performing
  unconstrained optimization of the top-$k$ score for the \emph{Bottlenose}
  dataset (top) and the \emph{Humpback} dataset (bottom).  Note how the weights
  shrink toward the edges.  For the \emph{Bottlenose} dataset the endpoint of
  the dorsal fin is often under water. For the \emph{Humpback} dataset
  the tips of the fluke are sometimes cropped, and the shortest-path algorithm
  often skips across the notch.
}
\vspace{-0.1in}
\label{fig:bernstein}
\end{figure}

We use the latter to initialize the search at $\mathbf{c} = \mathbf{1}$, which
leads to a uniform $\mathbf{w}$.  We set $n = 10$ and optimize for the
coefficients $\mathbf{c}$
using an open-source package for Bayesian optimization~\cite{Nogueira17}.
These coefficients $\mathbf{c}$ define a polynomial $f(x|\mathbf{c})$ on the interval $x \in [0,
1]$.  After defining $n$ uniformly-spaced points $x_1, x_2, \ldots, x_n$ on
this interval, where $n$ is the number of points on the trailing edge contour,
we compute the entries in the spatial weight vector as $w_i =
f(x_i|\mathbf{c})$.
This leads to the weight vectors shown in Figure~\ref{fig:bernstein}   for the
\emph{Bottlenose} and \emph{Humpback} datasets.  

\subsubsection{LNBNN Classification}
Similar to~\cite{Hughes16}, we use the local naive Bayes
nearest neighbor (LNBNN) algorithm~\cite{McCann12} to produce a
ranking of known individuals~\cite{Crall13}.  Our work differs
from~\cite{Hughes16} in that we use the integral curvature representation both
to compute descriptors and to determine keypoints.

\textbf{Feature Descriptor.}
Instead of using the Difference-of-Gaussian norm defined in Equation 1
in~\cite{Hughes16} to encode local shape information, we use the curvature
representation defined in Section~\ref{subsec:curvature-representation} of this
work.
Subsections of the curvature representation are resampled to a fixed length,
normalized by the Euclidean norm, and used as feature descriptors.

\textbf{Feature Keypoints.}
Similar to~\cite{Hughes16}, we choose the keypoints between which to define the subsections mentioned
above by resampling the curvature representation to a fixed length and choosing as
keypoints the $n - 2$ largest local extrema of the curvature representation at each
scale as well as the start and endpoints.  Combinations of these
keypoints yields ${n}\choose 2$ subsections per scale, between which we
extract the corresponding values from the curvature representation.

\textbf{LNBNN Classification.}
These feature descriptors are computed for all known
individuals in the database, and placed in a data structure for approximate
nearest neighbors using ANNOY~\cite{Bernhardsson13}.  We compute a score for
each individual using LNBNN classification, specifically Algorithm~2 as defined
in~\cite{McCann12}. The benefit of using LNBNN instead of standard approximate
nearest neighbors is that it considers not only the distance to the nearest
descriptor from a given individual, but also the distance to the nearest descriptor
from a different individual~\cite{Crall13}.  The difference between these is
used to update the score, which  reduces the contribution from non-distinctive
feature descriptors.

\section{Experiments}
\label{sec:experiments}
To demonstrate the effectiveness of our curvature representation and matching
algorithms, we evaluate the two approaches from Section~\ref{subsec:ranking}
for producing a ranking of known individuals on the \emph{Bottlenose} and
\emph{Humpback} datasets.  We use the top-$k$ score, defined as the fraction of
the time the correct individual appears in the first $k$ entries of the
ranking, to evaluate time-warping (with and without learned spatial
weights), as well as LNBNN using our curvature descriptor and the
Difference-of-Gaussian descriptor~\cite{Hughes16}.

In particular, we show that when using top-$1$ accuracy,
the curvature representation outperforms the Difference-of-Gaussian descriptor
by $95\%$ to $91\%$ on the \emph{Bottlenose}
dataset, and $80\%$ to $40\%$ on the \emph{Humpback} dataset.
To compare against Leafsnap~\cite{Kumar12}, we also construct the Histogram
of Curvature over Scale from our integral curvature representation, and
produce a ranking using the histogram intersection distance.  We were unable
to achieve good results with this, however, and we suspect that the reason is
that computing a histogram over the curvature representation loses the spatial
information of the nicks and notches necessary for individual identification.

Additionally, because humpback whales often have uniquely identifying patterns
of scarring and pigmentation on their flukes, we also compare against
HotSpotter~\cite{Crall13}, which uses LNBNN with SIFT
descriptors~\cite{Lowe04}.  The texture information captured by HotSpotter is
complementary to the curvature of the trailing edge, and so we also evaluate
combining our identification algorithms with HotSpotter.

We run experiments identical to those described in the following section on two
related datasets to ensure the generality of our approach.

\subsection{Defining Queries}
\label{subsec:defining-queries}
\textbf{Bottlenose Dolphins.} 
We randomly select $m = 10$ encounters for each individual, and use all the images
from these encounters for the database.  When an individual appears in only $n$
encounters such that $m > n$, we use $n-1$ encounters for the database so that
we have at least one query.  The images from remaining encounters
are used as query encounters.  We also investigate the effect of
varying $m$ on the top-$k$ accuracy in Section~\ref{subsec:num-encounters}.

\textbf{Humpback Whales.}
The \emph{Humpback} dataset typically contains only a single image per
individual in each encounter and two encounters per individual.  In practice,
this means that most individuals are represented by one image in the database.

When evaluating time-warping and HotSpotter, we use the minimum alignment cost across images
in the encounter as the similarity score.  For LNBNN, we stack the descriptors
from all images in the encounter to build the query.

We run all experiments on five random splits and report mean scores,
however, there is little variance across runs.

\subsection{Qualitative Results}
\begin{figure}[t]
\begin{center}
  \includegraphics[width=0.95\linewidth]{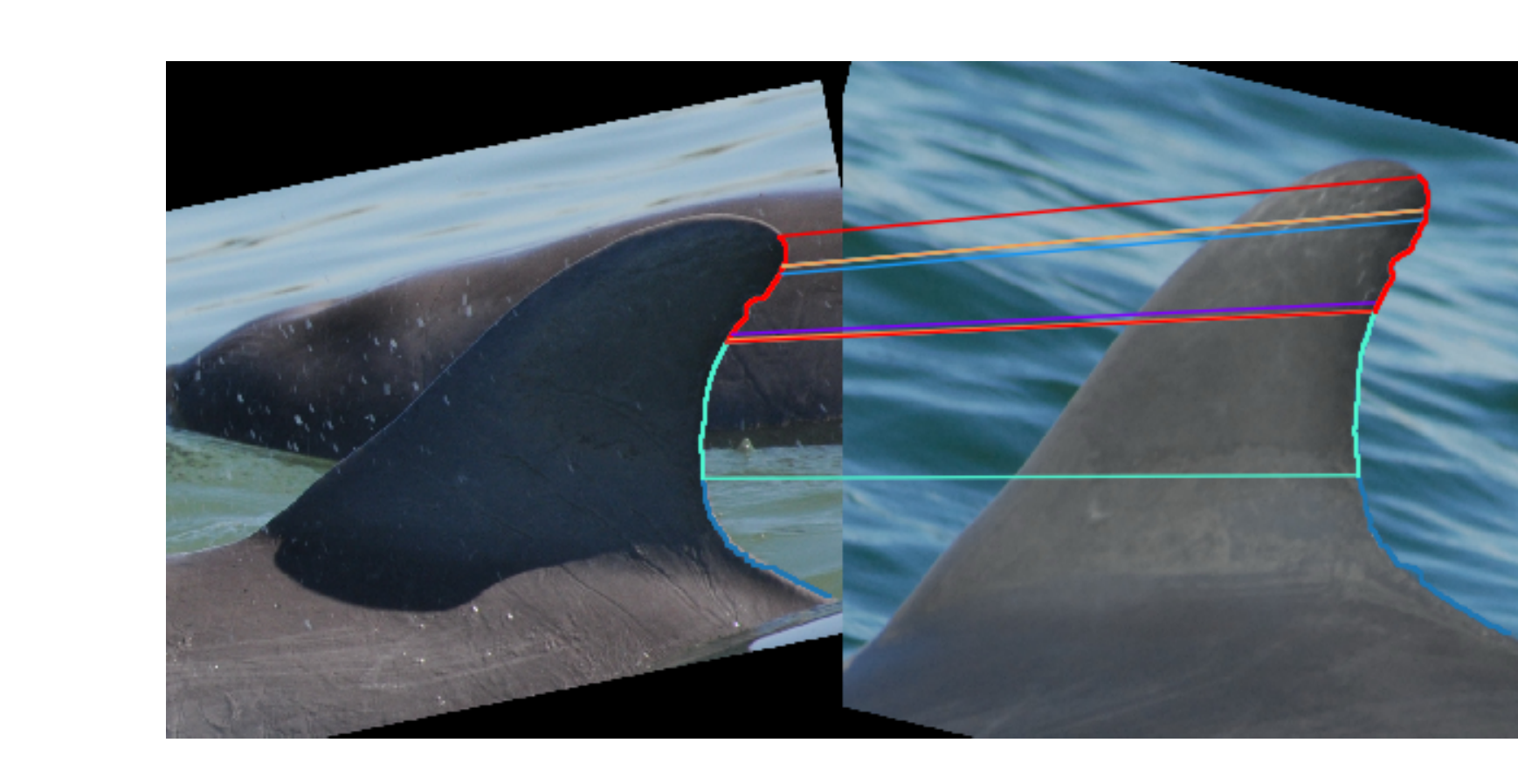}
  \includegraphics[width=0.95\linewidth]{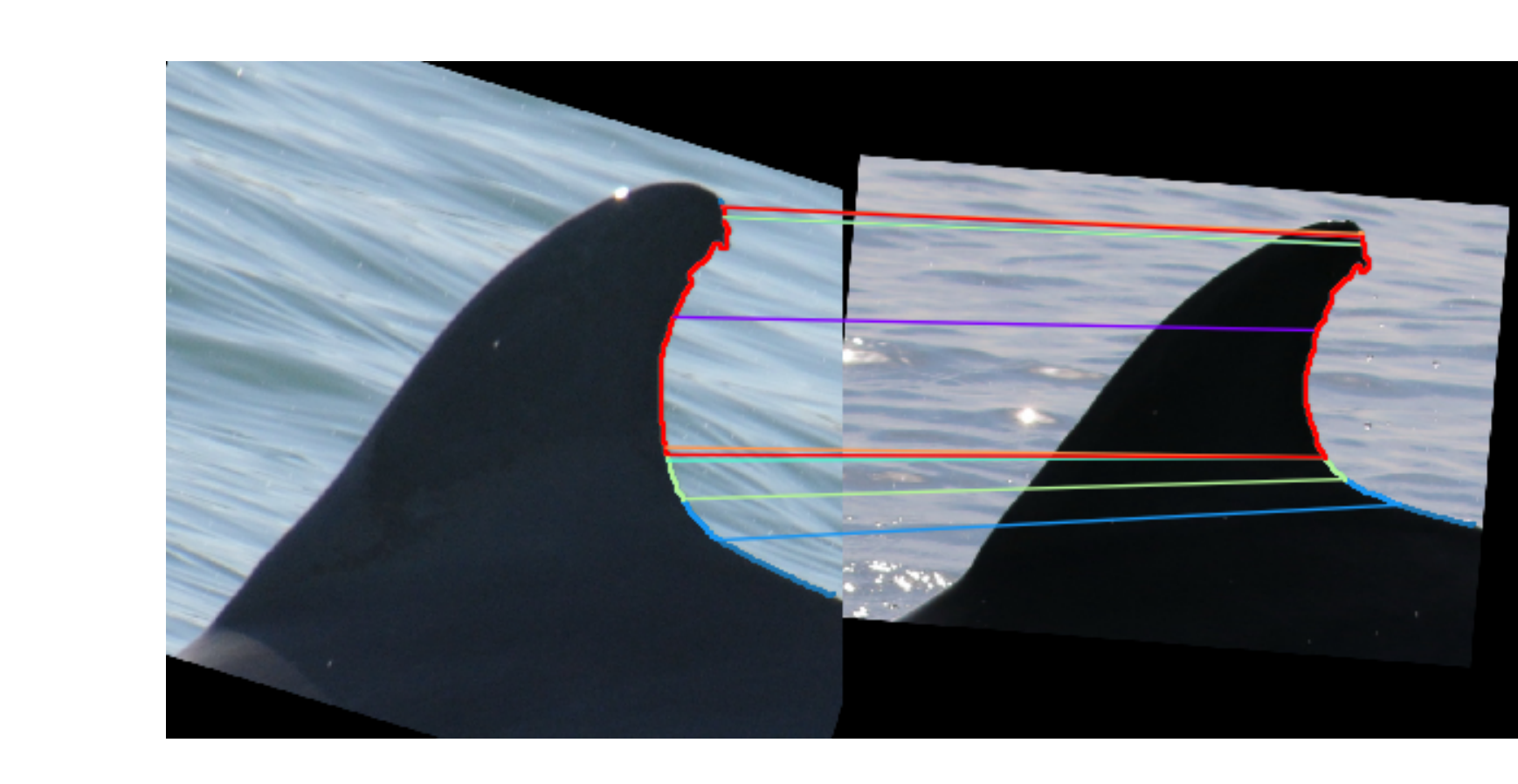}\\
  \includegraphics[width=0.95\linewidth]{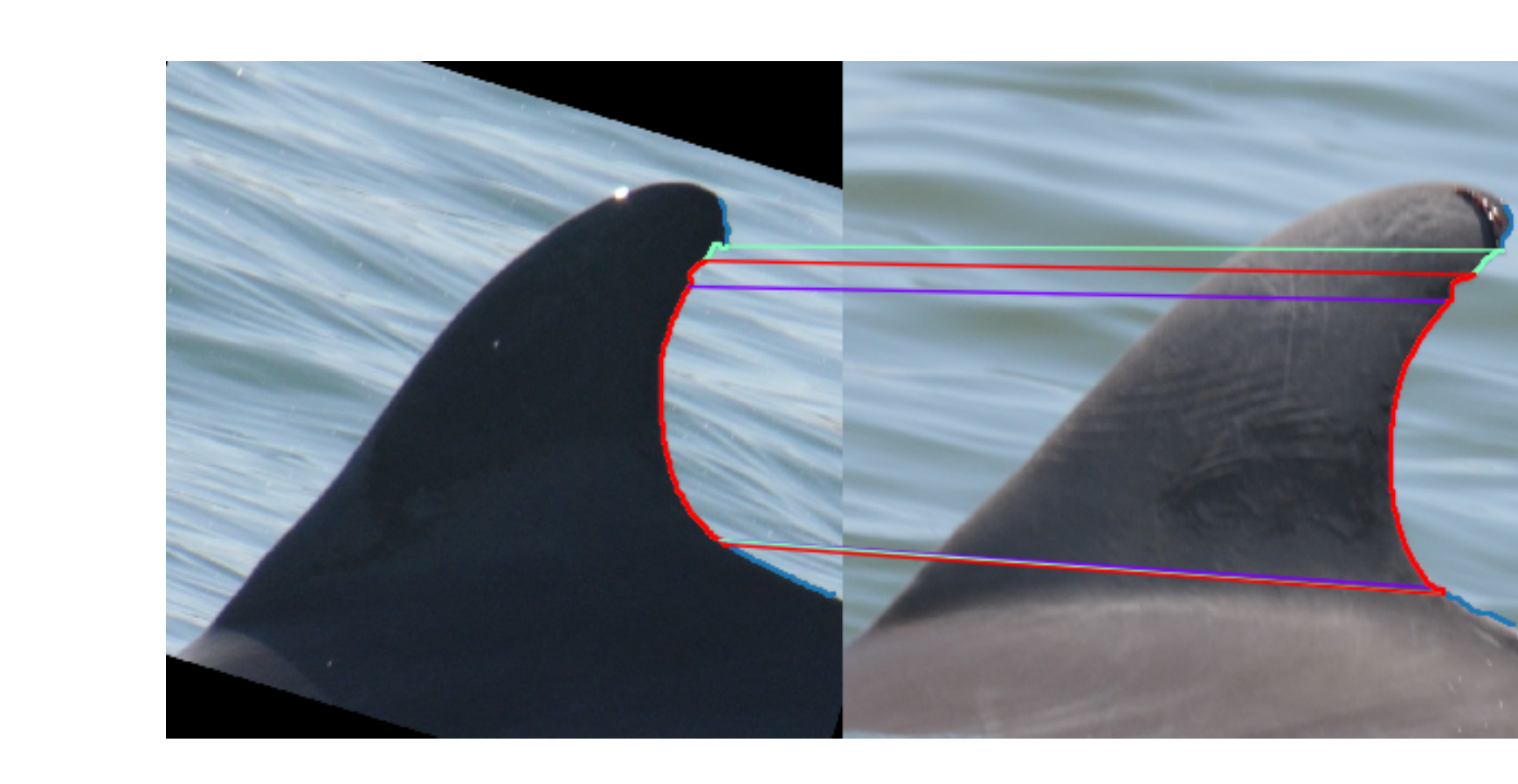}
\end{center}
\vspace{-0.1in}
\caption{\emph{Bottlenose} dataset.  The top row shows an instance where the
  correct individual is ranked first.  Note how the strongest match (shown in
  red) corresponds to the most distinct notch.  The middle row shows an
  instance where a different individual is ranked first, while the correct
  individual, ranked second, is shown in the bottom row.  For all rows, we show
  the query (left) and database (right) images that contribute most to the
  match score.  Weak LNBNN matches are not shown.
}
\vspace{-0.2in}
\label{fig:bottlenose-qual}
\end{figure}

\begin{figure}[t]
\begin{center}
  \includegraphics[width=0.475\linewidth]{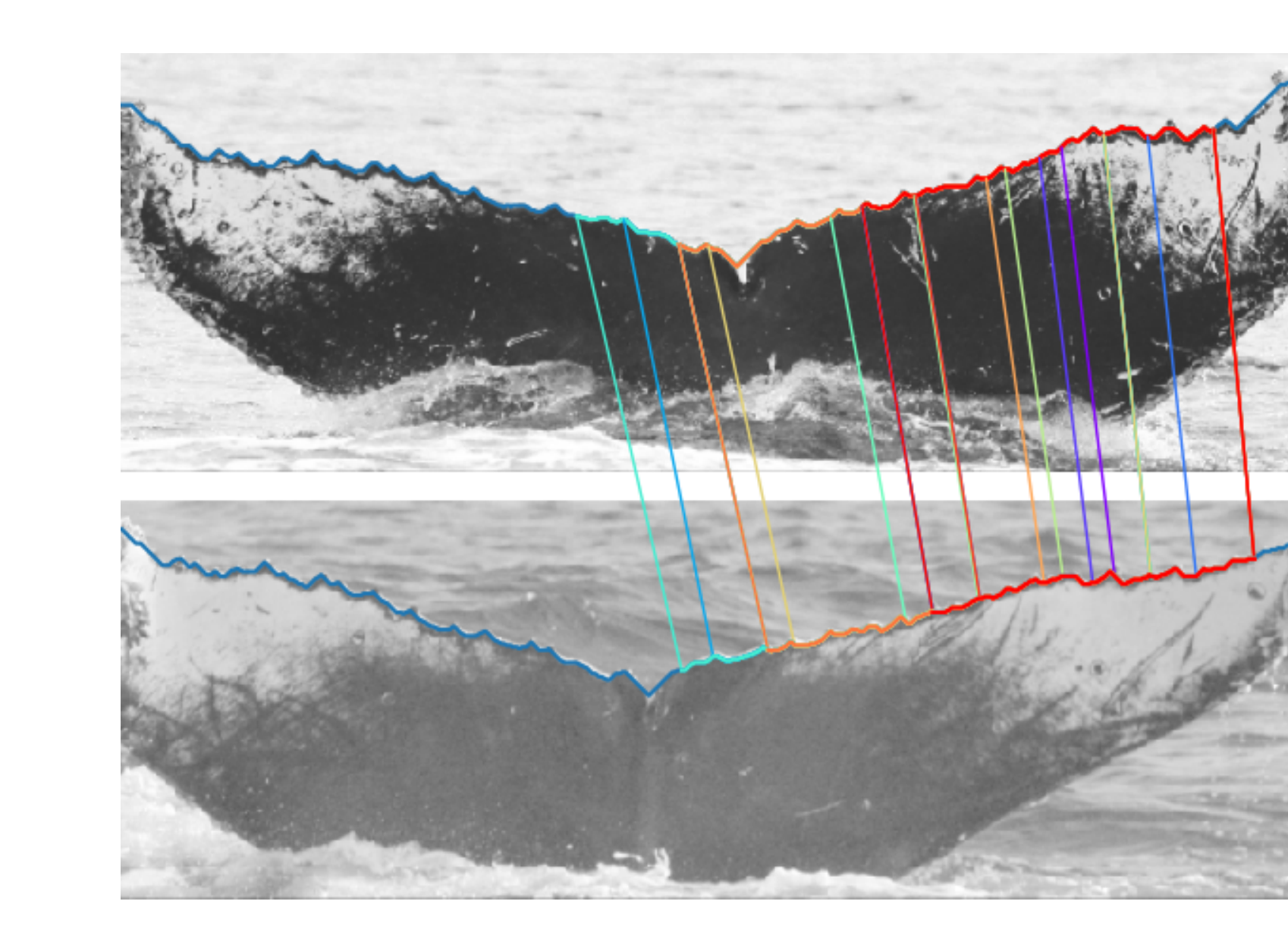}
  \includegraphics[width=0.475\linewidth]{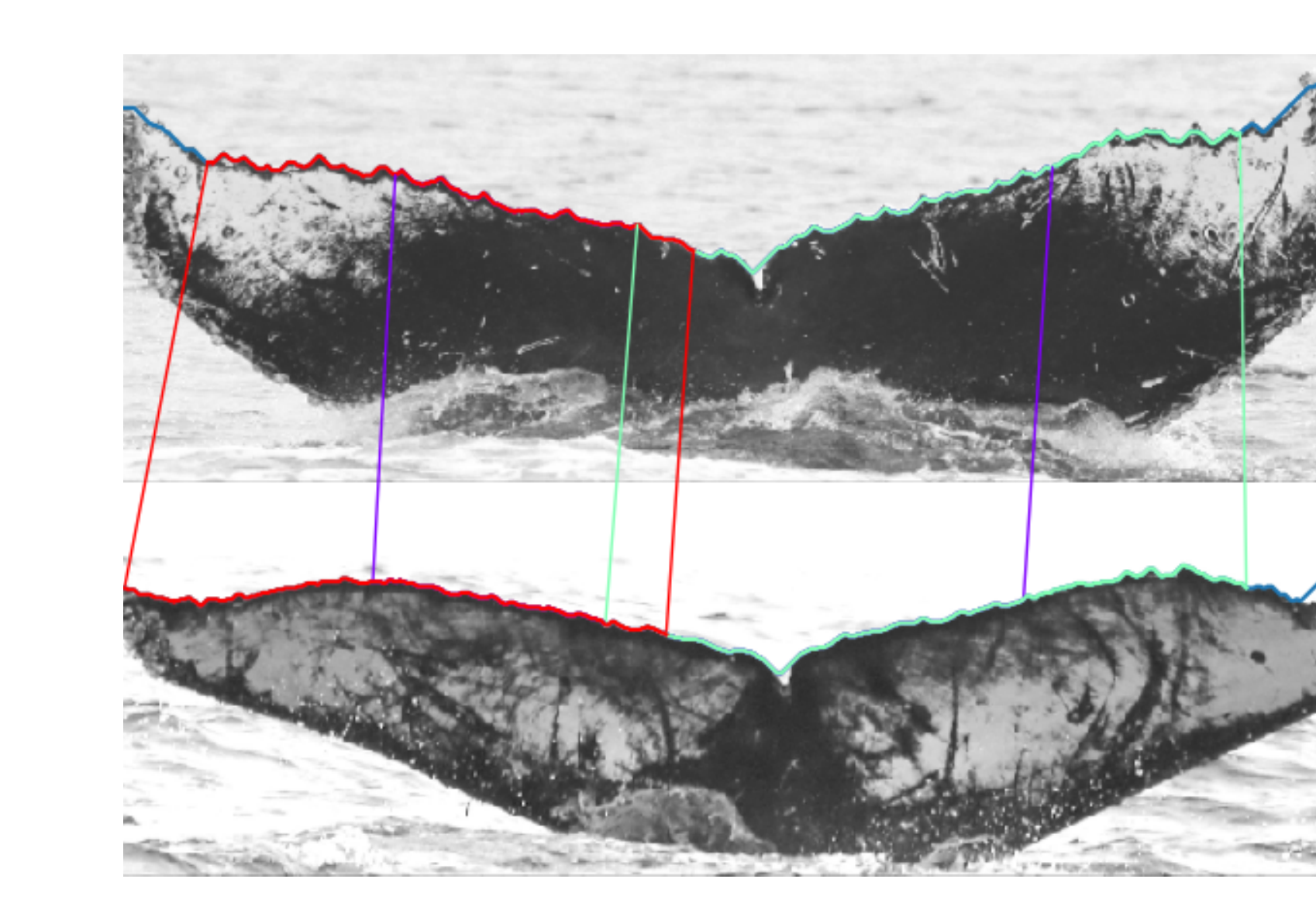}\\
  \includegraphics[width=0.475\linewidth]{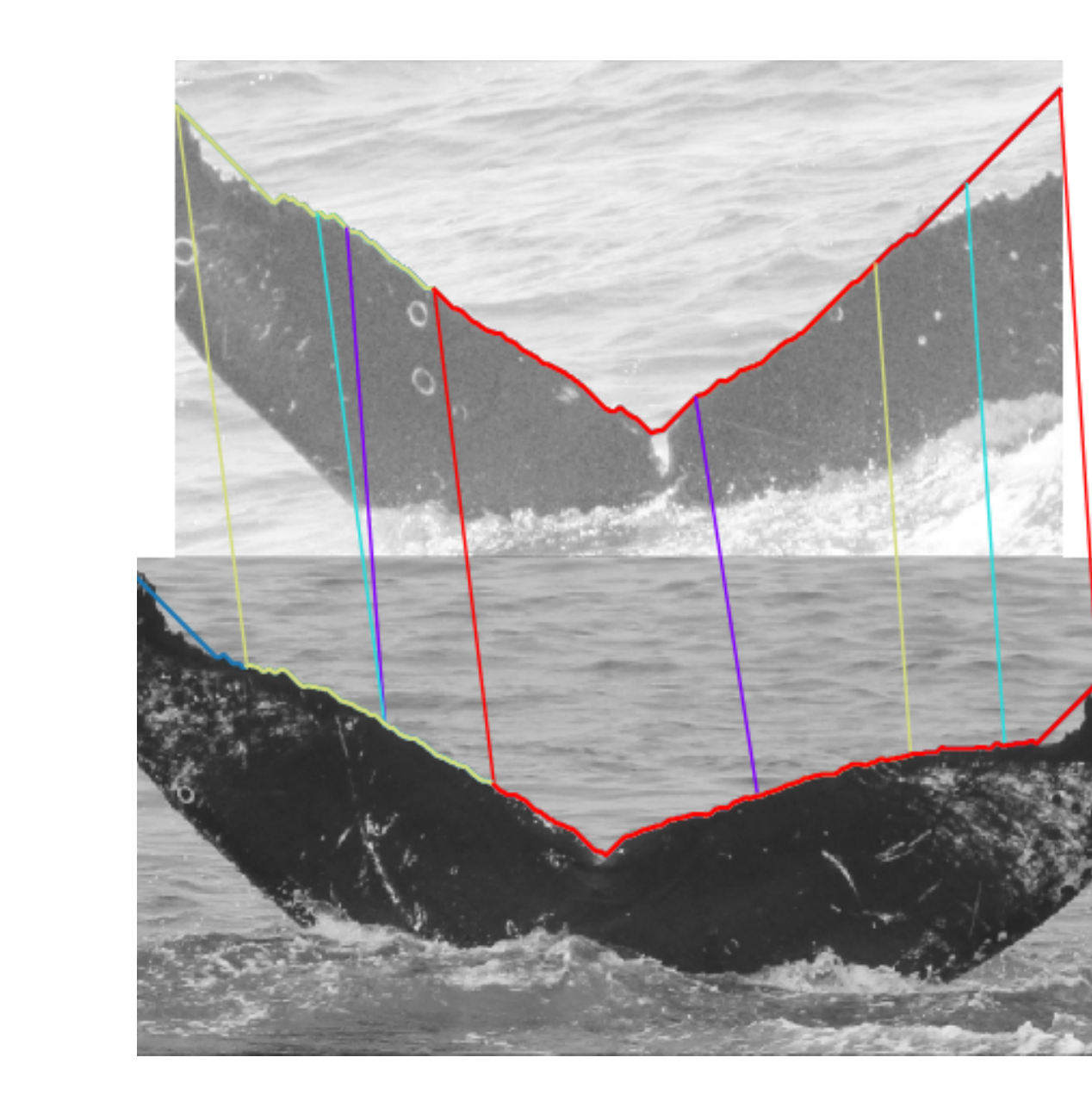}
  \includegraphics[width=0.475\linewidth]{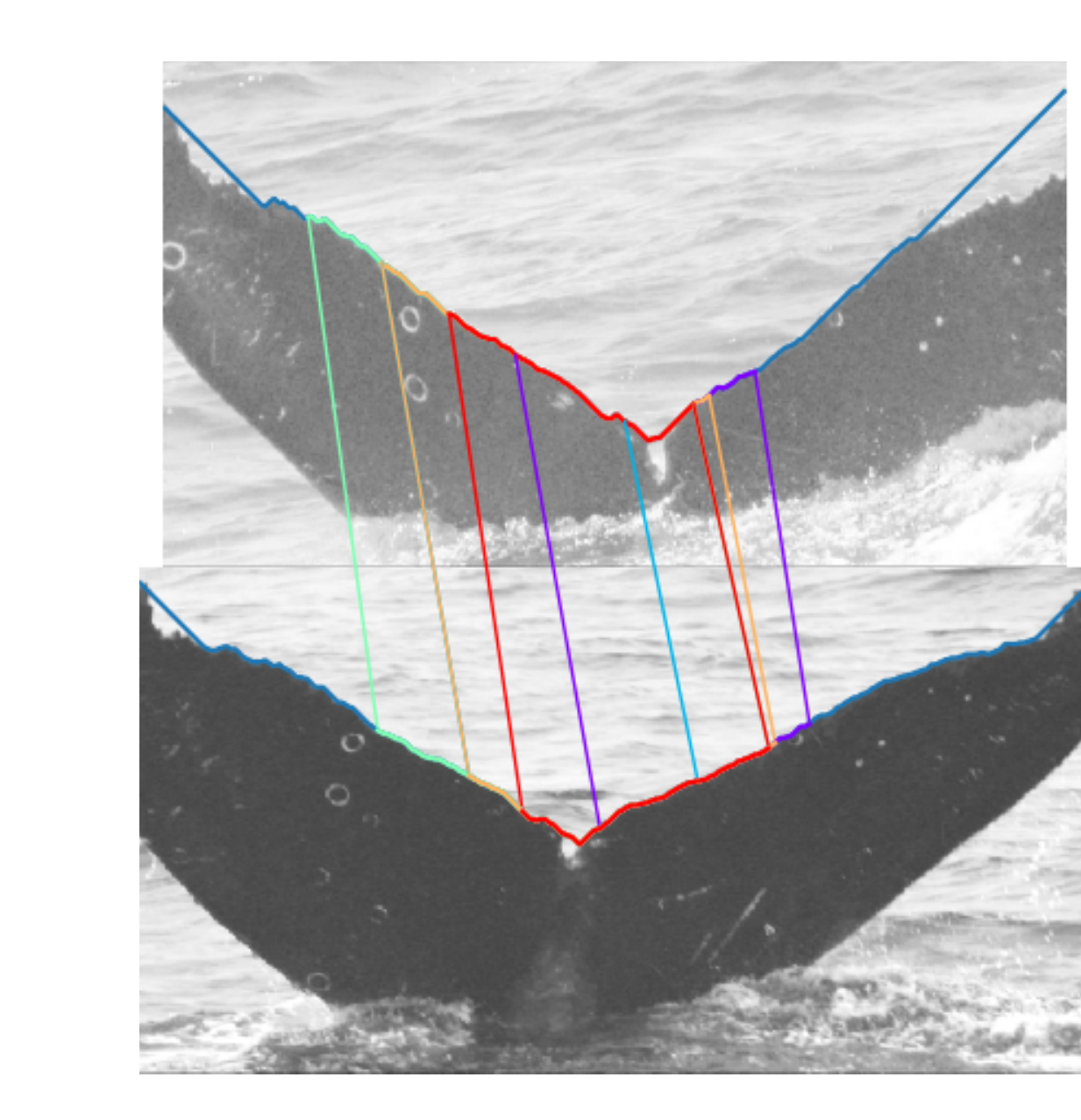}\\
\end{center}
\vspace{-0.1in}
\caption{\emph{Humpback} dataset.  The top four images show an instance where
  the correct individual is ranked first (top left), followed by the database
  individual ranked second (top right).  The bottom four images show an instance
  where a different individual is ranked first (bottom left), while the correct
  individual is ranked second (bottom right).  For both cases, we show the query (top)
  and database (bottom) images that contribute most to the total match score.
  Weak LNBNN matches are not shown.
}
\vspace{-0.15in}
\label{fig:humpback-qual}
\end{figure}

Before quantitatively evaluating
our algorithms, we show successful and unsuccessful identifications for
the \emph{Bottlenose} and \emph{Humpback} datasets in
Figures~\ref{fig:bottlenose-qual}~and~\ref{fig:humpback-qual}, respectively.
In each figure, we show the pair of images (query and database) that
contributes the most to the total score.  We plot a minimal subset of matches
such that the sum of the LNBNN scores from these matches is at least half the
total score.  Although the matches shown are sparse, in practice the entire
length of the contour is matched, albeit with lower scores.  Pairs of lines of
the same color indicate the start and endpoints of the trailing edge
corresponding to the matched curvature descriptor.  The matches are ordered
such that the strongest match is shown in red, and the weakest in purple.  The
sections of the trailing edge not covered by strong LNBNN matches is shown in
blue.

There are two main causes of misidentifications, namely (a) errors in the
contour extraction that cause distinguishing features to be poorly represented
in the curvature vectors, and (b)  distinctions between very smooth trailing
edges that are insufficiently valued by the matching algorithm.  Both are
amplified by significant viewpoint differences between database and query
trailing edges for the correct match. 



\subsection{Ranking Performance}
\label{subsec:ranking-performance}
To evaluate ranking performance, we compare two variations of each of
the algorithms from Section~\ref{subsec:ranking}.  Next, we
describe the parameter choices for each of these algorithms.

\textbf{Time-Warping Alignment.}
When using learned spatial weights,
we use the relevant $\mathbf{w}$ as shown in Figure~\ref{fig:bernstein} for the
\emph{Bottlenose} and \emph{Humpback} datasets.  For the \emph{Bottlenose}
dataset, we resample the curvature representation to $128$ points and set the
Sakoe-Chiba bound~\cite{Sakoe78} in the dynamic time-warping algorithm to $8$.
We use scales of $\{0.04, 0.06, 0.08, 0.10\}$.  For the
\emph{Humpback} dataset, these are set to $748$, $75$, and $\{0.02, 0.04, 0.06,
0.08\}$, respectively.  For both datasets, we determine the radii of the
circles used for integral curvature (Figure~\ref{fig:curvature}) by multiplying the scales by the maximum
dimension of the fin, specifically, the height for dorsal fins, and the width
for flukes.

\textbf{Descriptor Indexing.}
When doing LNBNN classification, we set the number of keypoints at which we
define contour subsections to $32$.  The keypoints are placed along the trailing edge
(which is resampled to $1024$ points) at
the points corresponding to the local extrema of the
curvature or Difference-of-Gaussian representation, as appropriate.
We set the dimension of the feature descriptors to $32$.  The
top-$k$ accuracy plateaus or slightly degrades on both datasets for larger
dimensions.  We speculate that this is due to the noisy nature of our extracted
trailing edges, where resampling to a smaller feature dimension acts as a form
of smoothing.  The scales for the Difference-of-Gaussian descriptors 
are the same as described in~\cite{Hughes16}.

\begin{figure}[t]
\begin{center}
  \includegraphics[width=0.9\linewidth]{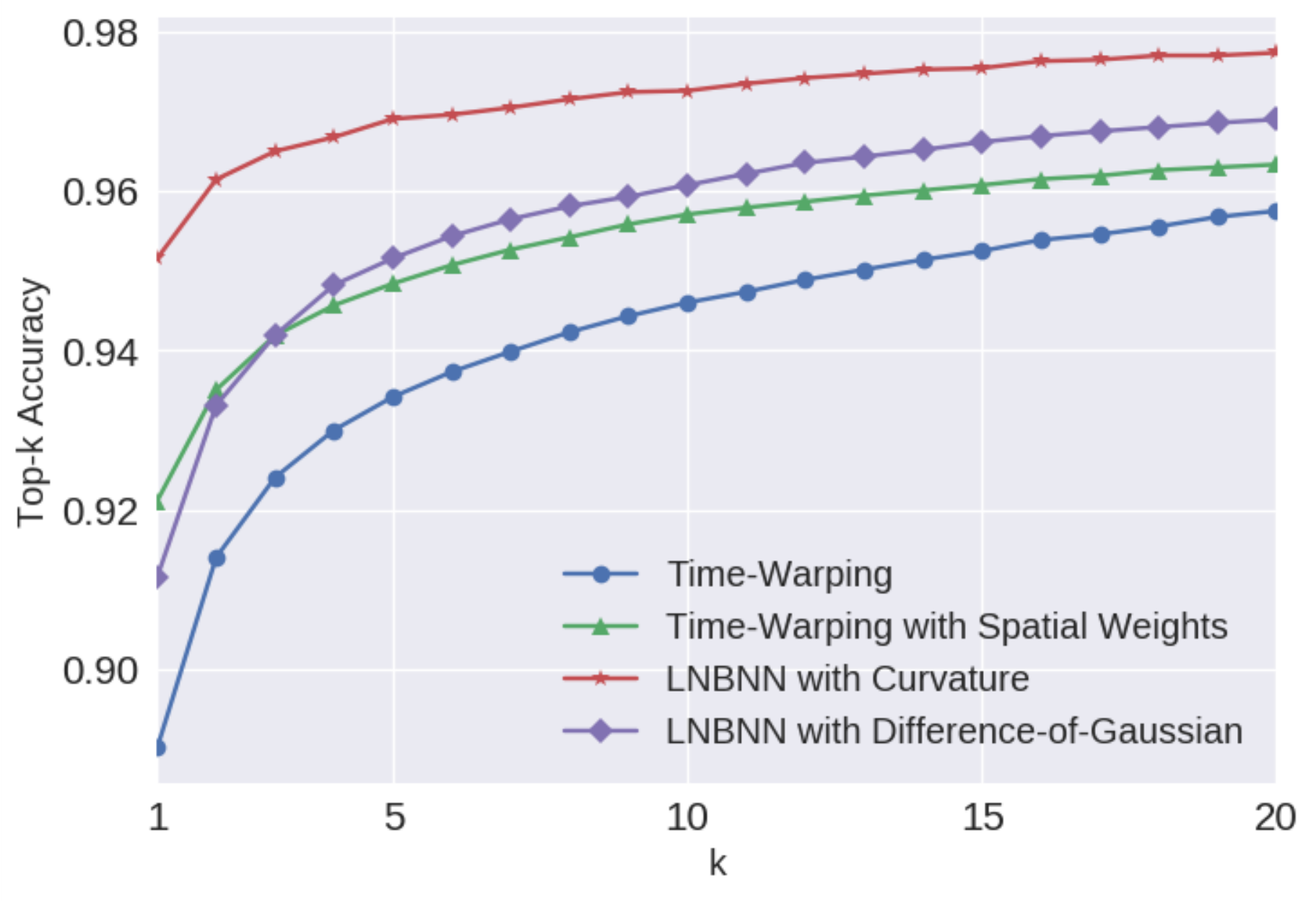}
\end{center}
\vspace{-0.1in}
 \caption{
   Top-$k$ scores for time-warping alignment and LNBNN for the
   \emph{Bottlenose} dataset.
}
\vspace{-0.2in}
\label{fig:bottlenose-topk}
\end{figure}

\begin{figure}[t]
\begin{center}
  \includegraphics[width=0.9\linewidth]{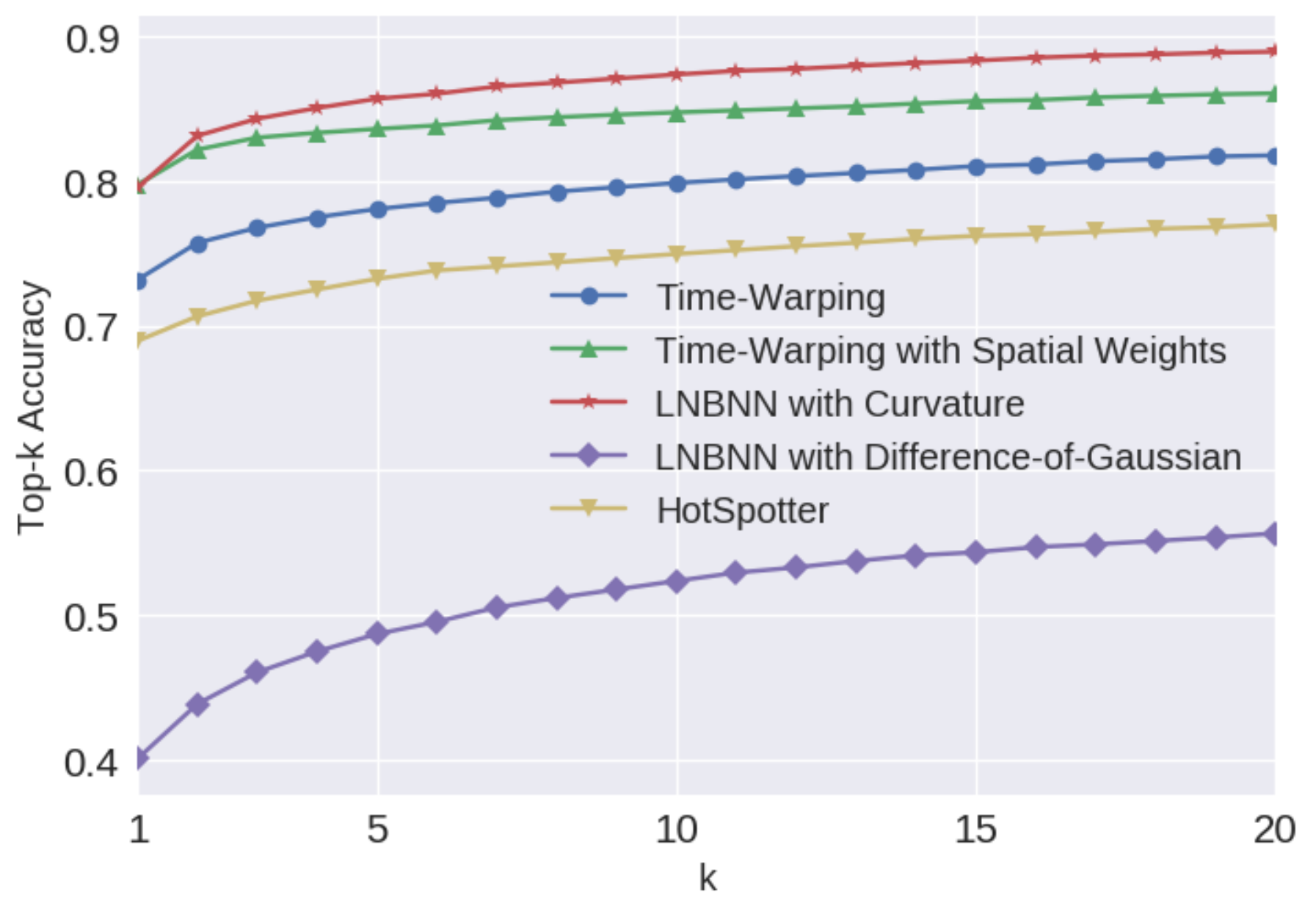}
\end{center}
\vspace{-0.1in}
 \caption{
   Top-$k$ scores for time-warping alignment, LNBNN, and HotSpotter for the
   \emph{Humpback} dataset.
}
\vspace{-0.1in}
\label{fig:humpback-topk}
\end{figure}

The results for the \emph{Bottlenose} and \emph{Humpback} datasets are
shown in Figures~\ref{fig:bottlenose-topk} and \ref{fig:humpback-topk},
respectively.  With LNBNN, the curvature descriptor outperforms the Difference-of-Gaussian
descriptor for both datasets.  We argue that this is because of the robustness
with which integral curvature can capture noisy local information --- with a
sufficiently large number of points representing the trailing edge, the
exact coordinates of any single point have little effect on the curvature
value.

\textbf{Evaluation.} The relative performance of the two matching algorithms is
different for the two datasets ---  it is likely that this is because of how
the identifying information is distributed.  For the \emph{Bottlenose} dataset,
where only a few distinct marks are useful for identification, the descriptor
indexing approach performs better.  For the \emph{Humpback} dataset, where the
information necessary for identification is spread along the entire length of
the trailing edge, the time-warping alignment approach achieves similar
results.

We repeat these experiments on two smaller datasets to determine if our
approach generalizes to datasets of the same (or similar) species.  On a common
dolphin dataset with $3744$ images representing $186$ individuals, we achieve
$74\%$ top-$1$ accuracy using time-warping and $69\%$ using LNBNN, and on a
humpback whale dataset with $1388$ images representing $419$ individuals, we
achieve $86\%$ top-$1$ accuracy using time-warping and $89\%$ using LNBNN.


\subsection{Number of Encounters per Known Individual}
\label{subsec:num-encounters}
\begin{figure}[t]
\begin{center}
  \includegraphics[width=0.9\linewidth]{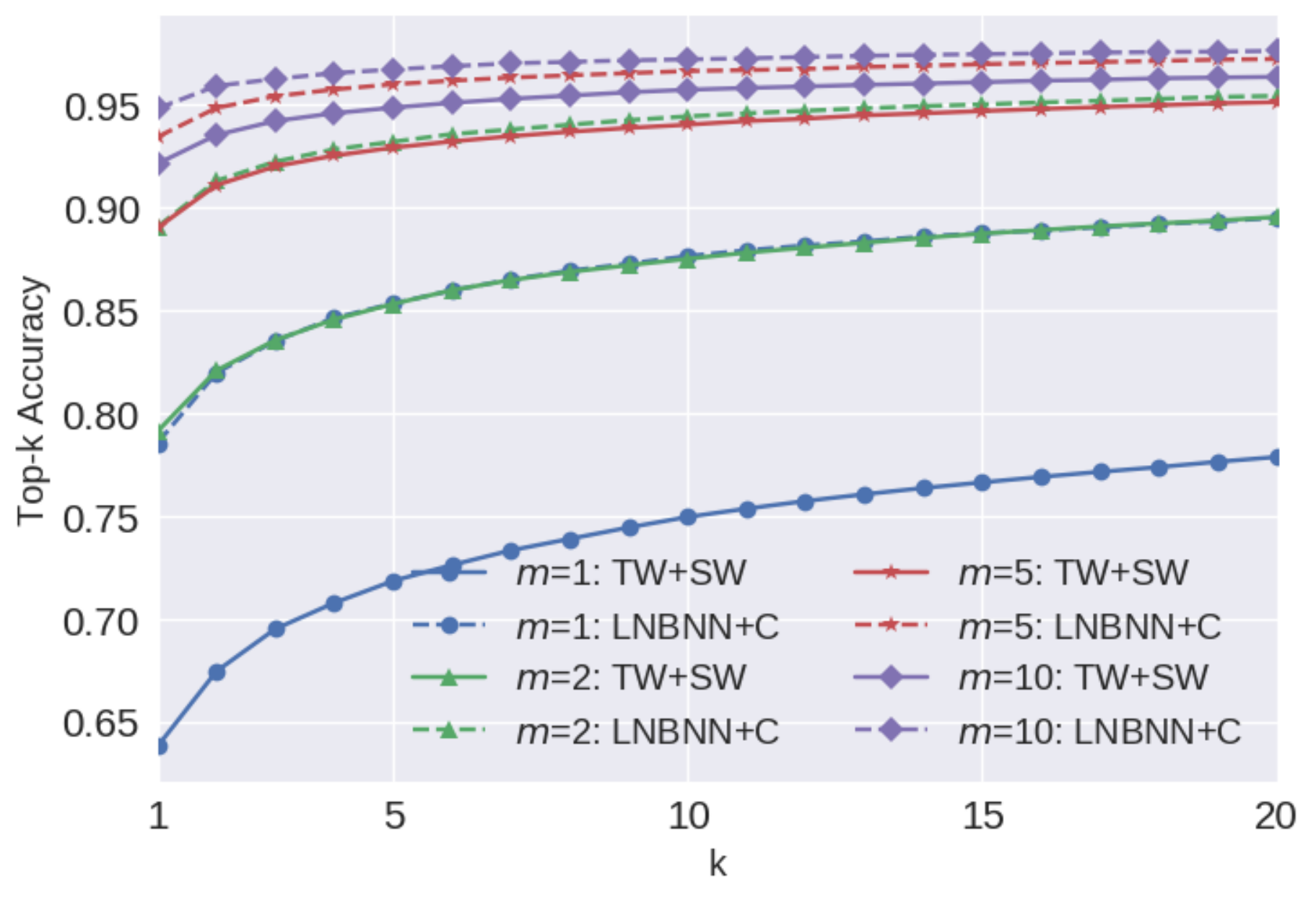}
\end{center}
\vspace{-0.1in}
 \caption{The effect of varying the number of encounters used to represent each
   individual in the database for the \emph{Bottlenose} dataset for the two
   best performing algorithms on this dataset, namely time-warping with spatial
   weights (TW+SW), as well as LNBNN with curvature descriptors (LNBNN+C).
}
\vspace{-0.1in}
\label{fig:bottlenose-num-encounters}
\end{figure}

While we choose the encounters from the \emph{Bottlenose} dataset randomly to
evaluate our algorithms, researchers may wish to choose the best images of each
known individual for the database.  Figure~\ref{fig:bottlenose-num-encounters}
shows the effect of increasing the number of encounters.  Adding more
encounters, and hence more images, to the database has several advantages.
First, there are more viewpoints represented, which makes it more likely that a
given query aligns well with one from the database.  Second, more
images per individual acts as insurance against the event where images may have
distinctive parts of the trailing edge occluded.  Choosing database images to
maximize the information content for identification is a problem we intend to
address in future work.

\subsection{Error Correlation}
\label{subsec:error-correlation}
\begin{figure}[t]
\begin{center}
  \includegraphics[width=0.9\linewidth]{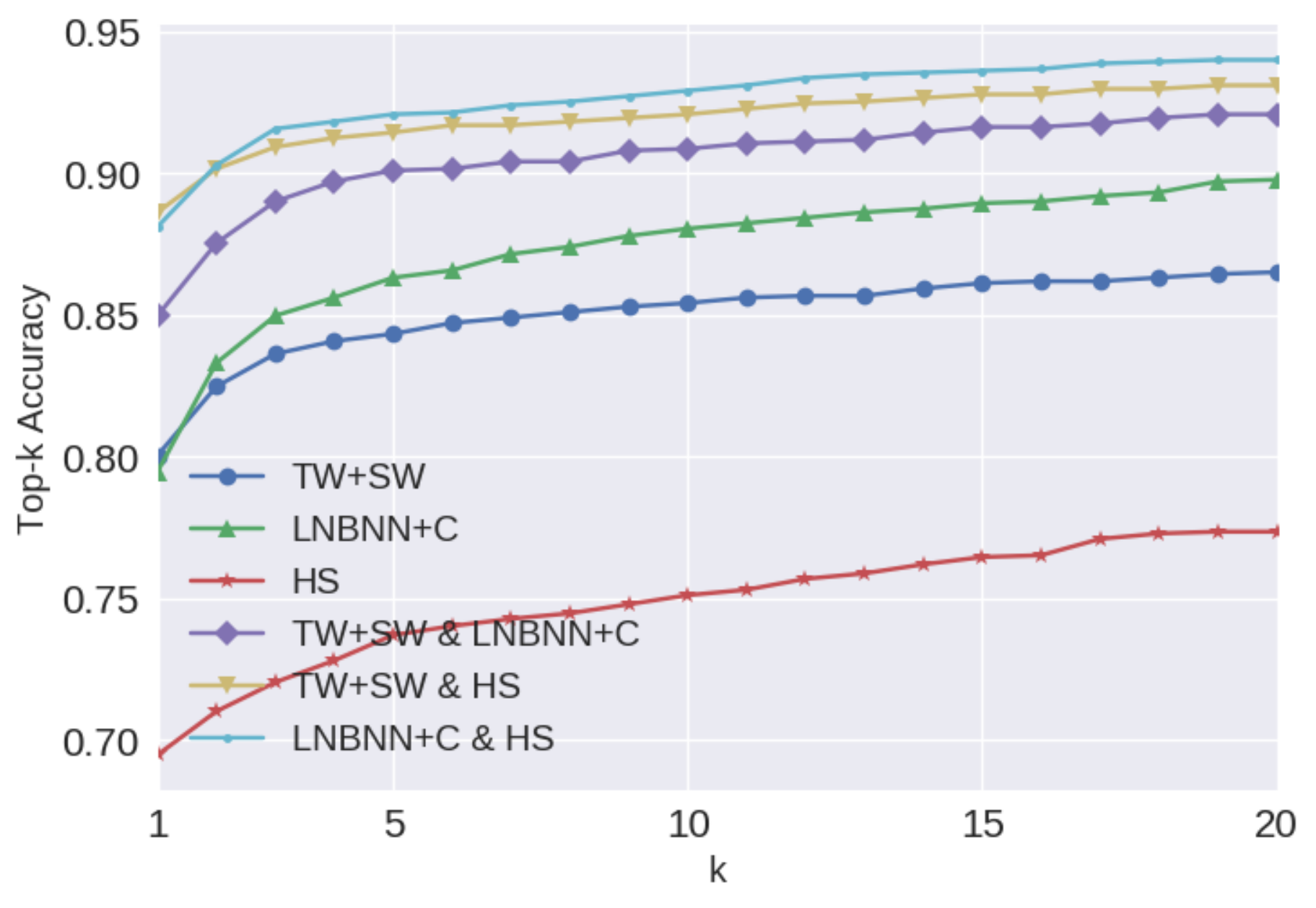}
\end{center}
\vspace{-0.1in}
 \caption{Combining algorithms that use complementary sources of information
 (fluke pigmentation and trailing edge curvature) improves performance on the
 \emph{Humpback} dataset.}
\vspace{-0.1in}
\label{fig:crc-error-correlation}
\end{figure}

In addition to evaluating the ranking performance of each algorithm separately,
we also explore the possibility of combining algorithms.  If the correct
individual appears in the top-$k$ entries of either of the algorithms, we
consider the match to be correct.  Combining our algorithms with
HotSpotter~\cite{Crall13} is
of particular interest to us, because of their complementary nature --- while
our matching algorithms use the integral curvature of the trailing edge,
HotSpotter uses SIFT~\cite{Lowe04} descriptors extracted from the interior of the fluke to
describe the unique patterns of pigmentation.
Figure~\ref{fig:crc-error-correlation} shows that augmenting time-warping
(TW+SW) with HotSpotter (HS) improves the top-$1$ accuracy from $80\%$ to
$89\%$, and augmenting LNBNN with HotSpotter improves the top-$1$ accuracy from
$79\%$ to $88\%$.

\section{Conclusion}
We introduced novel combinations of integral curvature representation and two
matching algorithms for identifying individual cetaceans from their fins.  This
representation captures the local pattern of nicks and notches in such a way
that they may be compared using either a time-warping algorithm or descriptor
indexing.  The effectiveness of our method is shown by computing accuracy
scores on two real-world datasets, each with distinct challenges.  For the
\emph{Bottlenose} dataset, with very little information per image, 
descriptor indexing outperforms time-warping because it considers not only the
feature distance, but also distinctiveness.  For the \emph{Humpback} dataset,
there are few images per individual, but many features per image.  The
time-warping algorithm is well-suited for this problem, because it preserves
the spatial integrity of curvature along the trailing edge while exploiting
learned spatial weights to emphasize matches from regions of stable curvature.
As a result, the performance of the two algorithms is similar.  In both
cases, we demonstrate that we achieve results that can greatly accelerate the
process of cetacean identification.

While the focus of this paper has not been on the details of the contour
extraction, a major focus of future work will be a unified algorithm that works
on both species and leads to a generalization that allows rapid adaptation to
new species. An important consideration will be to restrict the amount of
manually-generated training data required.

\pagebreak
{\small
\bibliographystyle{ieee}
\bibliography{dolphins}
}

\end{document}